\documentclass[runningheads]{llncs}
\usepackage{graphicx}

\usepackage{tikz}
\usepackage{comment}
\usepackage{amsmath,amssymb} 
\usepackage{color}
\usepackage{appendix}

\usepackage[accsupp]{axessibility}  

\usepackage{multirow}
\usepackage{graphicx}
\usepackage{subfigure} 
\usepackage[misc]{ifsym}
\usepackage{orcidlink}
\usepackage[capitalize]{cleveref}

\begin{document}
\pagestyle{headings}
\mainmatter
\def\ECCVSubNumber{52}  

\title{Understanding the Dynamics of DNNs Using Graph Modularity} 

\titlerunning{Graph Modularity}

\author{Yao Lu\inst{1}\orcidlink{0000-0003-0655-7814}
\and
Wen Yang\inst{1}\orcidlink{0000-0002-8525-5672} \and
Yunzhe Zhang\inst{1}\orcidlink{0000-0003-2662-608X} \and Zuohui Chen\inst{1}\orcidlink{0000-0003-1806-6676} \and Jinyin Chen\inst{1}\orcidlink{0000-0002-7153-2755} \and \\ Qi Xuan\inst{1} \textsuperscript{\Letter}\orcidlink{0000-0002-6320-7012} \and Zhen Wang\inst{2} \textsuperscript{\Letter}\orcidlink{0000-0002-8182-2852} \and Xiaoniu Yang\inst{1,3}\orcidlink{0000-0003-3117-2211}}
\authorrunning{Y. Lu et al.}

%
\institute{Institute of Cyberspace Security, Zhejiang University of Technology, Hangzhou, 310023. China\\ \email{\{yaolu.zjut,czuohui\}@gmail.com}, \email{\{chenjinyin,xuanqi\}@zjut.edu.cn}, \email{wenyang.zjut@outlook.com}, \email{xsgxlz@live.cn} \and School of Artificial Intelligence, Optics and Electronics (iOPEN), Northwestern Polytechnical University, Xi’an 710072. China\\ \email{zhenwang0@gmail.com} \and Science and Technology on Communication Information Security Control Laboratory, Jiaxing 314033, China\\ \email{yxn2117@126.com}} 
\maketitle

\begin{abstract}
    There are good arguments to support the claim that deep neural networks (DNNs) capture better feature representations than the previous hand-crafted feature engineering, which leads to a significant performance improvement. In this paper, we move a tiny step towards understanding the dynamics of feature representations over layers. Specifically, we model the process of class separation of intermediate representations in pre-trained DNNs as the evolution of communities in dynamic graphs. Then, we introduce modularity, a generic metric in graph theory, to quantify the evolution of communities. In the preliminary experiment, we find that modularity roughly tends to increase as the layer goes deeper and the degradation and plateau arise when the model complexity is great relative to the dataset. Through an asymptotic analysis, we prove that modularity can be broadly used for different applications. For example, modularity provides new insights to quantify the difference between feature representations. More crucially, we demonstrate that the degradation and plateau in modularity curves represent redundant layers in DNNs and can be pruned with minimal impact on performance, which provides theoretical guidance for layer pruning. Our code is available at \url{https://github.com/yaolu-zjut/Dynamic-Graphs-Construction}.
\end{abstract}

\begin{figure*}[hbpt]
  \centering
   \includegraphics[width=0.9\linewidth]{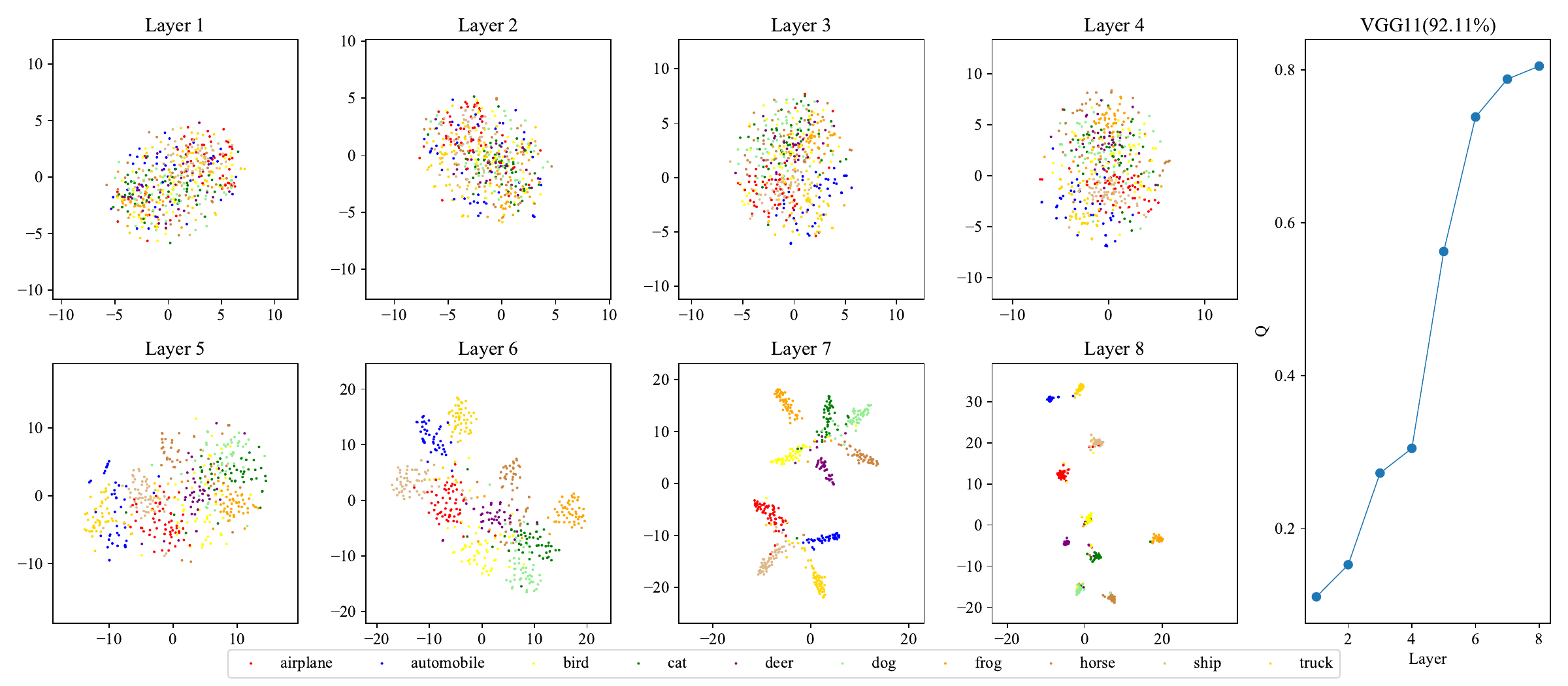}
   \caption{(Left): t-SNE outputs for CIFAR-10 testing data after each layer in VGG11. (Right): modularity curve of VGG11 on CIFAR-10. Best viewed in color.}
   \label{fig:t-SNE visualization}
\end{figure*}

\section{Introduction}
DNNs have gained remarkable achievements in many tasks, from computer vision \cite{he2016deep,redmon2016you,maqueda2018event} to natural language processing \cite{sutskever2014sequence}, which can arguably be attributed to powerful feature representations learned from data \cite{rumelhart1985learning,goh2021multimodal}. Giving insight into DNNs' feature representations is helpful to better understand neural network behavior, which attracts much attention in recent years. Some works seek to characterize feature representations by measuring similarities between the representations of various layers and various trained models~\cite{kornblith2019similarity,DBLP:conf/iclr/NguyenRK21,DBLP:conf/nips/RaghuGYS17,DBLP:conf/nips/MorcosRB18,DBLP:conf/nips/WangHGHW0H18,tang2020similarity,feng2020transferred}. Others visualize the feature representations in intermediate layers for an intuitive understanding, revealing that the feature representations in the shallow layers are relatively general, while those in the deep layers are more specific~\cite{das2020opportunities,DBLP:conf/eccv/ZeilerF14,selvaraju2017grad,wang2018visualizing,mahendran2015understanding}. These studies are insightful, but fundamentally limited, because they ignore the dynamics of DNNs or can only understand the dynamics of DNNs through qualitative visualization instead of quantitative study.

Hence, in this paper, we build upon previous studies and investigate the dynamics of intermediate layers. The left part of \cref{fig:t-SNE visualization} shows t-SNE \cite{van2008visualizing} outputs for 500 CIFAR-10 testing samples after each convolutional layer in VGG11, from which we are able to see how class separation in the feature representations progresses as the layer goes deeper. Inspired by this, we seek to quantify the process of class separation of intermediate representations for better understanding the dynamics of DNNs. Specifically, we treat each sample as a node, and there is an edge between two nodes if their feature representations are similar in the corresponding layer. Then we construct a series of graphs that share the same nodes, which can be modeled as a dynamic graph due to the feature continuity. In this way, we convert quantifying the process of class separation of intermediate representations to investigate the evolution of communities in dynamic graphs. Then, we introduce modularity to quantify the evolution of communities. As shown in the right part of \cref{fig:t-SNE visualization}, the value of modularity indeed grows with the depth, which is consistent with the process of class separation shown in the left part of \cref{fig:t-SNE visualization}. This indicates that modularity provides a quantifiable interpretation perspective for understanding the dynamics of DNNs.

Then we conduct systematic experiments on exploring how the modularity changes in different scenarios, including the training process, standard and adversarial scenarios. Through further analysis of the modularity, we provide two application scenarios for it: (i) representing the difference of various layers. (ii) providing theoretical guidance for layer pruning. To summarize, we make the following contributions:
\begin{itemize}
\item[$\bullet$] We model the class separation of feature representations from layer to layer as the evolution of communities in dynamic graphs, which provides a novel perspective for researchers to better understand the dynamics of DNNs.
\item[$\bullet$] We leverage modularity to quantify the evolution of communities in dynamic graphs, which tends to increase as the layer goes deeper, but descends or reaches a plateau at particular layers. To preserve the generality of modularity, systematic experiments are conducted in various scenarios, e.g., standard scenarios, adversarial scenarios and training processes.
\item[$\bullet$] Additional experiments show that modularity can also be utilized to represent the difference of various layers, which can provide insights on further theoretical analysis and empirical studies.
\item[$\bullet$] Through further analysis on the degradation and plateau in the modularity curves, we demonstrate that the degradation and plateau reveal the redundancy of DNNs in depth, which provides a theoretical guideline for layer pruning. Extensive experiments show that layer pruning guided by modularity can achieve a considerable acceleration ratio with minimal impact on performance. 
\end{itemize}

\section{Related Work}
\label{sec:related work}
Many researchers have proposed various techniques to analyze certain aspects of DNNs. Hence, in this section, we would like to provide a brief survey of the literature related to our current work.

\textbf{Understanding feature representations.} Understanding feature representations of a DNN can obtain more information about the interaction between machine learning algorithms and data than the loss function value alone. Previous works on understanding feature representations can be mainly divided into two categories. One category quantitatively calculates the similarities between the feature representations of different layers and models \cite{kornblith2019similarity,DBLP:conf/iclr/NguyenRK21,DBLP:conf/nips/RaghuGYS17,DBLP:conf/nips/MorcosRB18,DBLP:conf/nips/WangHGHW0H18,tang2020similarity,feng2020transferred}. For example, Kornblith et al.
\cite{kornblith2019similarity} introduce centered kernel alignment (CKA) to measure the relationship between intermediate representations. Feng et al. \cite{feng2020transferred} propose a new metric, termed as transferred discrepancy, to quantify the difference between two representations based on their downstream-task performance. Compared to previous studies which only utilize feature vectors, Tang et al. \cite{tang2020similarity} leverage both feature vectors and gradients into designing the representations of DNNs. On the basis of \cite{kornblith2019similarity}, Nguyen et al. \cite{DBLP:conf/iclr/NguyenRK21} utilize CKA to explore how varying depth and width affects model feature representations, and find that overparameterized models exhibit the block structure. Through further analysis, they show that some layers exhibit the block structure and can be pruned with minimal impact on performance. Another category attempts to obtain an insightful understanding of feature representations through interpreting feature semantics~\cite{das2020opportunities,DBLP:conf/eccv/ZeilerF14,selvaraju2017grad,wang2018visualizing,mahendran2015understanding}. Wang et al. \cite{wang2018visualizing} and Zeiler et al. \cite{DBLP:conf/eccv/ZeilerF14} discover the hierarchical nature of the features in the neural networks. Specifically, shallow layers extract basic and general features while deep layers learn more specifically and globally. Yosinski et al.~\cite{DBLP:conf/nips/YosinskiCBL14} quantify the degree to which a particular layer is general or specific. Besides, Donahue et al.~\cite{donahue2014decaf}  investigate the transfer of feature representations from the last few layers of a DNN to a novel generic task. 

\textbf{Modularity and community in DNNs.} Previous empirical studies have explored modularity and community in neural networks. Some seek to investigate learned modularity and community structure at the neuron level\cite{watanabe2018modular,watanabe2019understanding,watanabe2019interpreting,davis2020network,hod2021detecting,you2020graph} or at the subnetwork level \cite{csordas2021neural,lake2017building}. Others train an explicitly modular architecture \cite{alet2018modular,DBLP:conf/iclr/GoyalLHSLBS21} or promote modularity via parameter isolation \cite{DBLP:conf/nips/KirschKB18} or regularization \cite{delange2021continual} during training to develop more modular neural networks. Different from these existing works, in this paper, we explore the evolution of communities at the feature representation level, which provides a new perspective to characterize the dynamics of DNNs.

\textbf{Layer pruning.} State-of-the-art DNNs often involve very deep networks, which are bound to bring massive parameters and floating-point operations. Therefore, many efforts have been made to design compact models. Pruning is one stream among them, which can be roughly devided into three categories, namely, weight pruning~\cite{DBLP:conf/iclr/FrankleC19,azarian2020learned,DBLP:conf/iclr/LinSBDJ20}, filter pruning~\cite{lin2019towards,lin2020hrank,DBLP:conf/cvpr/Tang0XD0T021} and layer pruning~\cite{xu2020layer,elkerdawy2020filter,zhou2021evolutionary,chen2018shallowing,wang2021accelerate,wang2019dbp}. Weight pruning compresses over-parameterized models by dropping redundant individual weights, which has limited applications on general-purpose hardware. Filter pruning seeks to remove entire redundant filters or channels instead of individual weights. Compared to weight pruning and filter pruning, layer pruning removes the entire redundant layers, which is more suitable for general-purpose hardware. Existing layer pruning methods mainly differ in how to determine which layers need to be pruned. For example, Xu et al.~\cite{xu2020layer} first introduce a trainable layer scaling factor to identify the redundant layers during the sparse training. And then they prune the layers with very small factors and retrain the model to recover the accuracy. Elkerdawy et al. \cite{elkerdawy2020filter} leverage imprinting to calculate a per-layer importance score in one-shot and then prune the least important layers and fine-tune the shallower model. Zhou et al.~\cite{zhou2021evolutionary} leverage the ensemble view of block-wise DNNs and employ the multi-objective optimization paradigm to prune redundant blocks while avoiding performance degradation. Based on the observations of~\cite{alain2016understanding}, Chen and Zhao~\cite{chen2018shallowing}, Wang et al.~\cite{wang2021accelerate}, and Wang et al.~\cite{wang2019dbp}, respectively, utilize linear classifier probes to guide the layer pruning. Specifically, they prune the layers which provide minor contributions on boosting the performance of features. 

\textbf{Adversarial samples.} Although DNNs have gained remarkable achievements in many tasks \cite{he2016deep,redmon2016you,maqueda2018event,sutskever2014sequence}, they have been found vulnerable to adversarial examples, which are born of intentionally perturbing benign samples in a human-imperceptible fashion~\cite{DBLP:journals/corr/SzegedyZSBEGF13,DBLP:journals/corr/GoodfellowSS14}. 
The vulnerability to adversarial examples hinders DNNs from being applied in safety-critical environments. Therefore, attacks and defenses on adversarial examples have attracted significant attention in machine learning. There has been a multitude of work studying methods to obtain adversarial examples~\cite{DBLP:journals/corr/GoodfellowSS14,DBLP:conf/iclr/MadryMSTV18,papernot2016limitations,carlini2017towards,su2019one,narodytska2016simple,maho2021surfree,li2021aha} and to make DNNs robust against adversarial examples~\cite{shafahi2019adversarial,DBLP:conf/iclr/WongRK20}.

\begin{figure}[!t]
  \centering
   \includegraphics[width=0.9\linewidth]{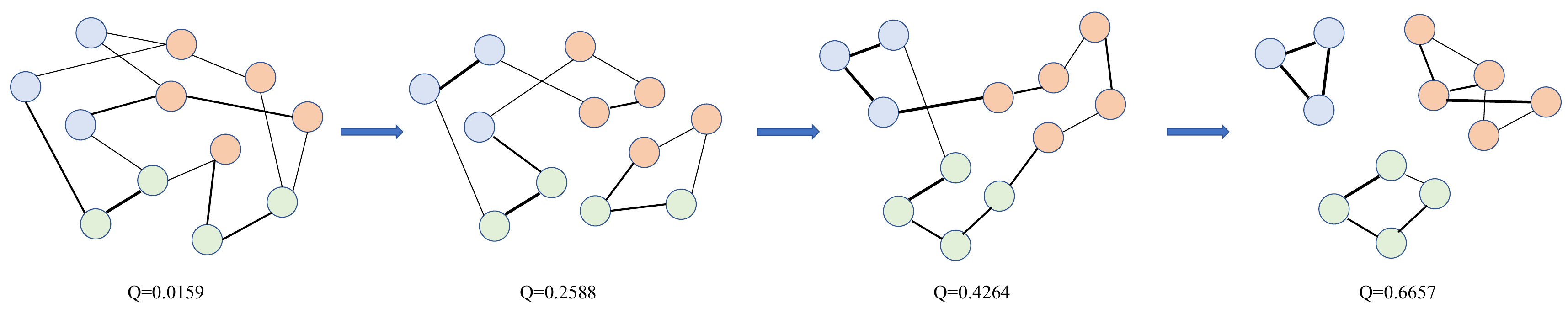}
   \caption{A dynamic graph with $4$ snapshots. Nodes of the same color represent that they are in the same community and the thickness of the line represents the weight of the edge. we use \cref{eq:important} to calculate the modularity.}
   \label{fig:demo}
\end{figure}

\section{Methodology}
\label{sec:methodology}
\subsection{Preliminary}
\label{preliminary}
We start by introducing some concepts in graph theory. 

\textbf{Dynamic graphs} are the graphs that change with time~\cite{harary1997dynamic}. In this paper, we focus on edge-dynamic graphs, i.e., edges may be added or deleted from the graph. Given a dynamic graph: $\mathcal{DG} = \{ \mathit{G}_1, \mathit{G}_2, \cdots, \mathit{G}_T \}$, where $T$ is the number of snapshots. Repeatedly leveraging the static methods on each snapshot can collectively look insight into the graph’s dynamics. \textbf{Communities}, which are quite common in many real networks~\cite{zachary1977information,lusseau2003emergent,jonsson2006cluster}, are defined as sets of nodes that are more densely connected internally than externally~\cite{newman2006modularity,newman2004finding}. \textbf{Ground-truth communities} can be defined as the sets of nodes with common properties, e.g., common attribute, affiliation, role, or function \cite{yang2015defining}. \textbf{Modularity}, as a common measurement to quantify the quality of communities \cite{DBLP:journals/corr/abs-0906-0612,newman2004finding}, carries advantages including intelligibility and adaptivity. In this paper, we adopt modularity $Q$ as following: 
\begin{equation}
  Q = \frac{1}{2W} \mathit{\sum}_{ij} \left( \mathit{a}_{ij} - \frac{\mathit{s}_i\mathit{s}_j}{2W}\right) \delta(\mathit{c}_i,\mathit{c}_j),
  \label{eq:important}
\end{equation} 
where $\mathit{a}_{ij}$ is the weight of the edge between node $i$ and node $j$, $W = \mathit{\sum}_i \mathit{\sum}_j \mathit{a}_{ij}$ is the sum of the weights of all edges, which is used as a normalization factor. $\mathit{s}_i = \mathit{\sum}_j{\mathit{a}_{ij}}$ and $\mathit{s}_j = \mathit{\sum}_i{\mathit{a}_{ij}}$ are the strength of nodes $i$ and $j$, $c_i$ and $c_j$ denote the community that nodes $i$ and $j$ belong to, respectively. $\delta(\mathit{c}_i,\mathit{c}_j)$ is 1 if node $i$ and node $j$ are in the same community and 0 otherwise. In this paper, each node represents an image with the corresponding label. Hence, these nodes can be divided into the corresponding ground-truth communities, which we utilize to calculate the modularity. 
\cref{fig:demo} gives an example to intuitively understand the evolution of communities, from which we find that the communities with a high value of modularity tend to strengthen the intra-community connections and weaken the inter-community connections.

\begin{figure*}[!t]
  \centering
   \includegraphics[width=0.75\linewidth]{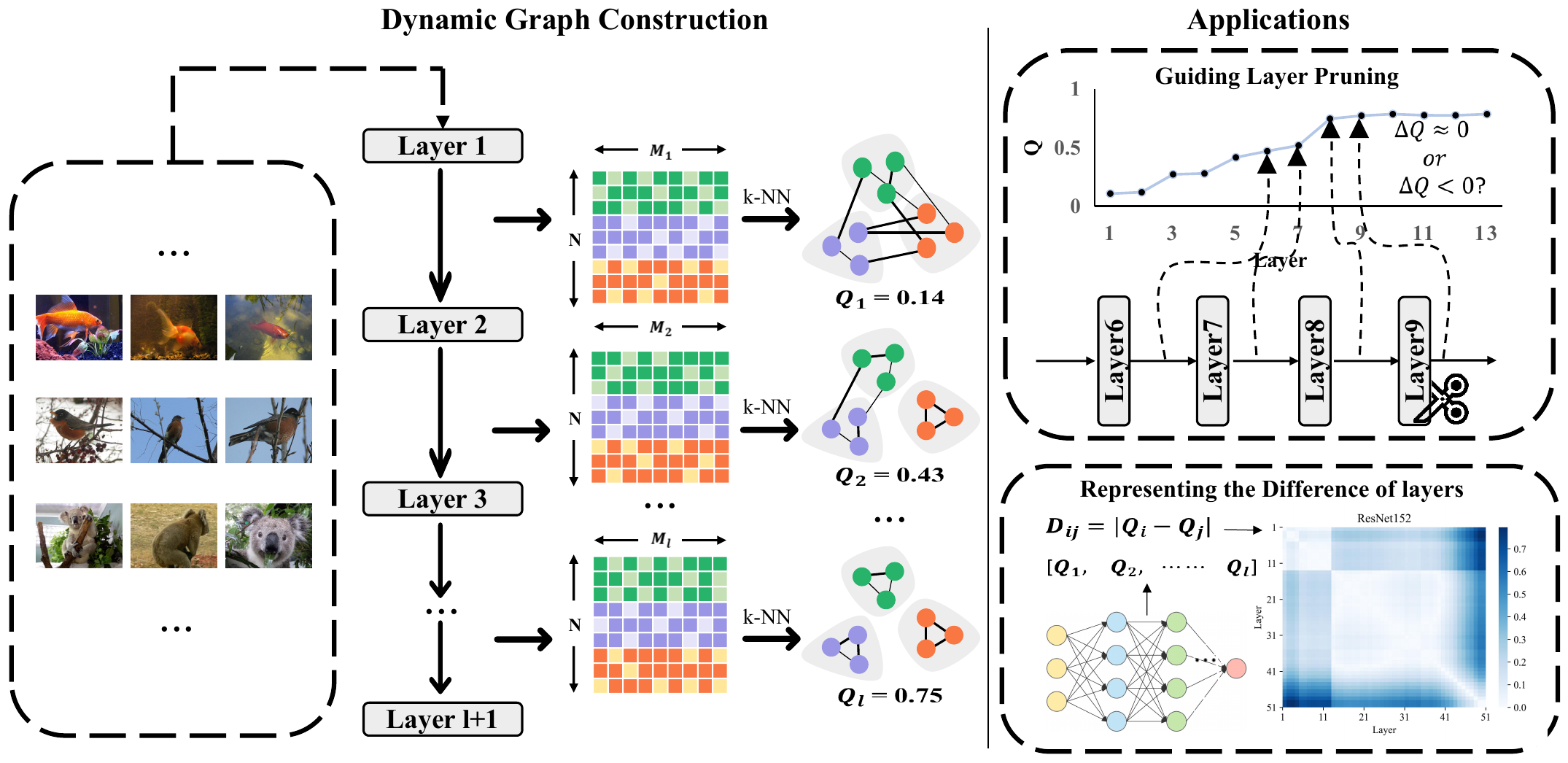}
   \caption{Pipeline for the dynamic graph construction and the application scenarios of the modularity metric. Best viewed in color, zoomed in for details.}
   \label{fig:arch}
\end{figure*}

\subsection{Dynamic Graph Construction}
\label{Graph Construction}
Our proposed dynamic graph construction framework to understand the dynamics of a given DNN is visually summarized in \cref{fig:arch}.

Considering a typical image classification problem, the feature representation of the sample is transformed over the course of many layers, to be finally utilized by a classifier at the last layer. Therefore, we can model this process as follows: 
\begin{equation}
    \tilde{y} = f_{l+1}(f_l(\dots f_1(x) \dots)),
  \label{eq:classification}
\end{equation} 
where $f_1(\cdot)$, $f_2(\cdot)$, $\cdots$, $f_l(\cdot)$ are $n$ functions to extract feature representations of layers from bottom to top, $f_{l+1}(\cdot)$ and $l$ denote the classifier and the number of layers, respectively. In this paper, we treat a bottleneck building block or a building block as a layer for ResNets \cite{he2016deep} and take a sequence of consecutive layers, e.g., Conv-BN-ReLu, as a layer for VGGs \cite{DBLP:journals/corr/SimonyanZ14a}. We randomly sample $N$ samples $\mathcal{X} = \{ \mathit{x}_1, \mathit{x}_2, \dots , \mathit{x}_N \}$ with corresponding labels $\mathcal{Y} = \{ \mathit{y}_1, \mathit{y}_2, \dots , \mathit{y}_N \}$ from the test set and feed them into a well-optimized DNN with fixed parameters to obtain an intermediate representation set $\mathcal{R} = \{r_1, r_2, \cdots, r_l\}$, where $r_i \in \mathbb{R}^{N \times \mathit{C}_i \times \mathit{W}_i \times \mathit{H}_i}$ denotes the feature representations in $i$-th layer, $\mathit{C}_i$, $\mathit{W}_i$ and $\mathit{H}_i$ are the number of channels, width and height of feature maps in $i$-th layer, respectively. Then we apply a flatten mapping $\mathit{f}: \mathbb{R}^{N \times \mathit{C} \times \mathit{W} \times \mathit{H}} \rightarrow \mathbb{R}^{N \times \mathit{M}}$ to each element in $\mathcal{R}$, where $\mathit{M} = \mathit{C} \times \mathit{W} \times \mathit{H}$. In order to capture the underlying relationship of samples in feature space, we construct a series of k-nearest neighbor (k-NN) graphs $\mathit{G}_{i} = (\mathit{A}_{i}, \mathit{r}_i)$, where $\mathit{A}_{i}$ is the adjacency matrix of the k-NN graph in $i$-th layer. Specifically, we calculate the similarity matrix $S = \mathbb{R}^{N \times N}$ among $N$ nodes using \cref{eq:cos}.\begin{equation}
\mathit{S}_{jk}^i = 
\begin{cases}
\dfrac{{\mathit{r}_{i,j}}^T \mathit{r}_{i,k}}{||\mathit{r}_{i,j}||\ ||\mathit{r}_{i,k}||},\ &\text{if}\ j \neq k \\
0,\ &\text{if}\ j = k \\
\end{cases},
\label{eq:cos}
\end{equation} 
where $\mathit{r}_{i,j}$ and $\mathit{r}_{i,k}$ are the feature representation vectors of samples $j$ and $k$ in $i$-th layer. According to the obtained similarity matrix $\mathit{S}_{jk}^i$, we choose top \textbf{k} similar node pairs for each node to set edges and corresponding weights, so as to obtain the adjacency matrix $\mathit{A}_{i}$. Now, we obtain a series of k-NN graphs $\{ \mathit{G}_{i} = (\mathit{A}_{i}, r_i) | i = 1, 2, \cdots, l\}$, which reveal the internal relationship between feature representations of different samples in various layers. Due to the continuity of feature representations, i.e., feature representation of the current layer is obtained on the basis of the previous one. Hence, these k-NN graphs are relevant to each other and can be treated as multiple snapshots of the dynamic graph $\mathcal{DG} = \{ \mathit{G}_1, \mathit{G}_2, \cdots, \mathit{G}_l \}$ at different time intervals.

Then we revisit the dynamic graph for a better understanding of community evolution. \cref{fig:demo} exhibits a demo, from which we can intuitively understand the process of community evolution in the dynamic graph. $\mathcal{DG}$ consists of $l$ snapshots, each snapshot shares same nodes (samples) and reveals the inherent correlations between samples in the corresponding layer. Therefore, our dynamic graph is actually an edge-dynamic graph. Due to the existence of ground-truth label of each sample, we can easily divide samples into $K$ ground-truth communities. Hence, we can calculate the modularity of each snapshot in $\mathcal{DG}$ using \cref{eq:important} together with the ground-truth communities. According to the obtained modularity of each snapshot, we finally obtain a modularity set $\mathcal{Q} = \{ \mathit{Q}_1, \mathit{Q}_2, \cdots, \mathit{Q}_l \}$, which reveals the evolution of communities in the dynamic graph.

\section{Experiments}
\label{experiments}
Our goal is to intuitively understand the dynamics in well-optimized DNNs. Reflecting this, our experimental setup consists of a family of VGGs \cite{DBLP:journals/corr/SimonyanZ14a} and ResNets \cite{he2016deep} trained on standard image classification datasets CIFAR-10 \cite{krizhevsky2009learning}, CIFAR-100 \cite{krizhevsky2009learning} and ImageNet \cite{DBLP:journals/ijcv/RussakovskyDSKS15} (For the sake of simplicity, we choose 50 classes in original ImageNet, termed as ImageNet50). Specifically, we leverage stochastic gradient descent algorithm with an initial learning rate of 0.01 to optimize the model. The batch size, weight decay, epoch and momentum are set to 256, 0.005, 150 and 0.9, respectively. The statistics of pre-trained models and ImageNet50 are shown in Appendix B. All experiments are conducted on two NVIDIA Tesla A100 GPUs. If not special specified, we set $k = 3$, $N = 500$ for CIFAR-10 and ImageNet50, $k = 3$, $N = 1000$ for CIFAR-100 to construct dynamic graphs.

\begin{figure}[htbp]
\centering
 \subfigure[VGGs]
 {
  \begin{minipage}[c]{0.499\textwidth}
   \centering
   \includegraphics[width=0.99\textwidth]{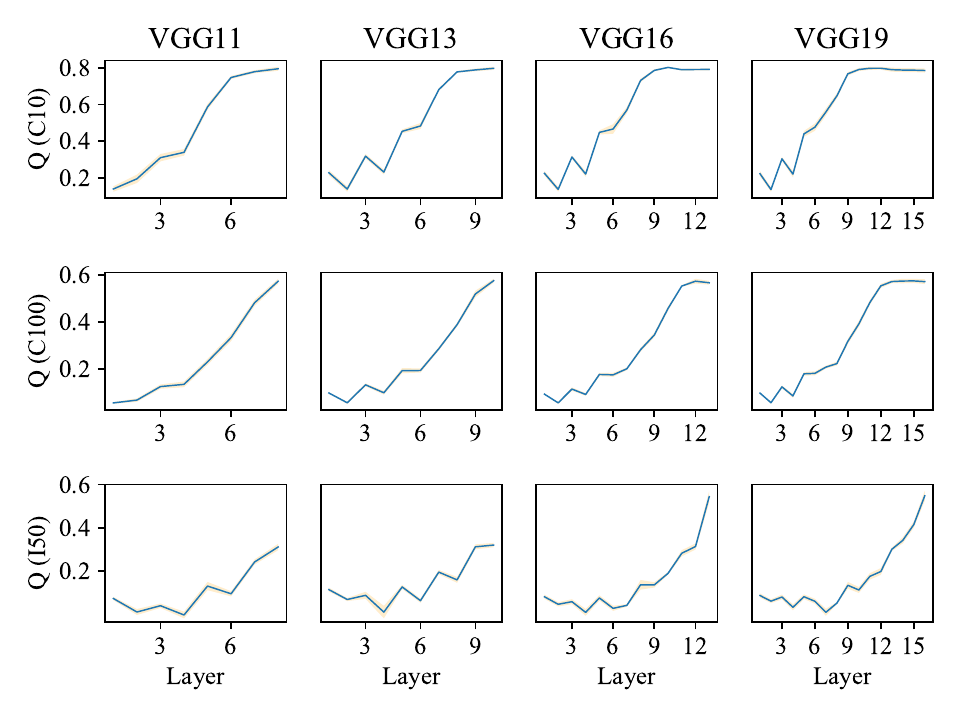}
  \end{minipage}%
  }%
 \subfigure[ResNets]
  {
  \begin{minipage}[c]{0.499\textwidth}
   \centering
   \includegraphics[width=0.99\textwidth]{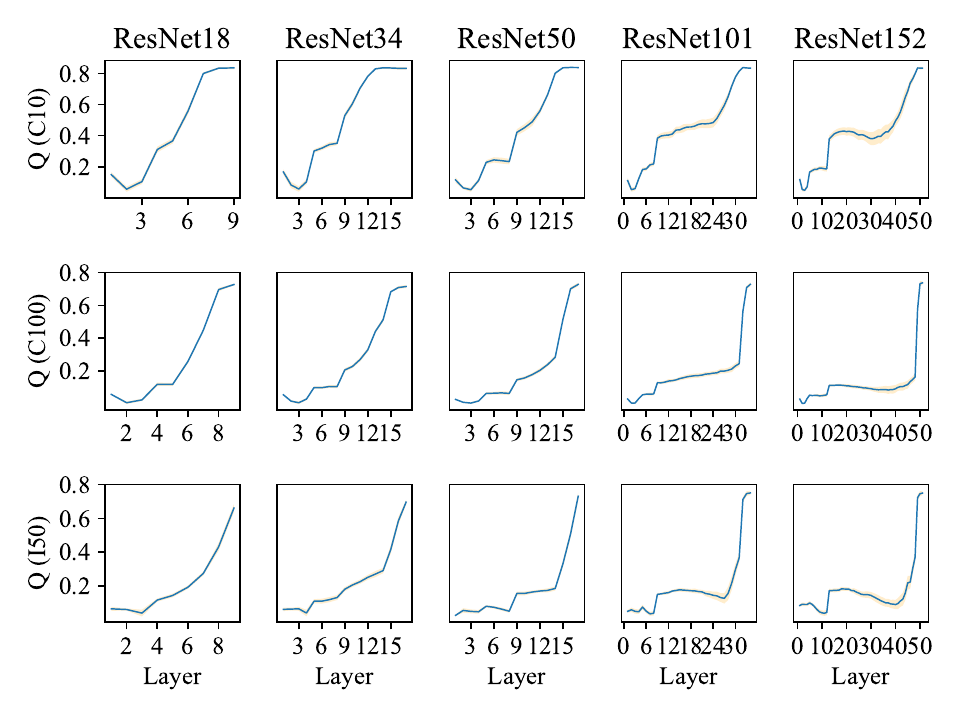}
  \end{minipage}
  }
\caption{Modularity curves of different models. C10, C100 and I50 denote CIFAR-10, CIFAR-100 and ImageNet50, respectively. Shaded regions indicate standard deviation.}
\label{modularity:VGGs and ResNets}
\end{figure}

\begin{figure*}[!t]
  \centering
   \includegraphics[width=0.9\linewidth]{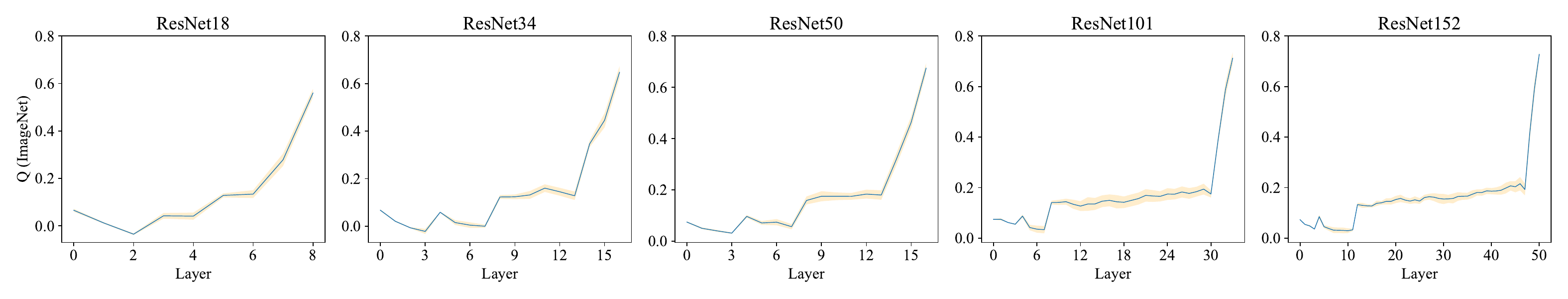}
   \caption{Modularity curves of ResNets on ImageNet.}
   \label{fig:ImageNet_modularity.pdf}
\end{figure*}

\subsection{Understanding the Evolution of Communities}
\label{Understanding the Evolution of Communities}
We systematically investigate the modularity curves of various models and repeat each experiment $5$ times to obtain the mean and variance of modularity curves. From the results reported in \cref{modularity:VGGs and ResNets}, we have the following observations:

\begin{itemize}
\item[$\bullet$] The modularity roughly tends to increase as the layer goes deeper. 

\item[$\bullet$] On the same dataset, the frequency of degradation and plateau existing in the modularity curve gradually increases as models get deeper. Specifically, the modularity curve gradually reaches a plateau on CIFAR-10 and CIFAR-100 at the deep layer as VGG gets deeper. Compared to VGGs, the modularity curve of ResNets descends, reaches a plateau, or rises very slowly mostly happening in the repeatable layers. 
\item[$\bullet$] According to the modularity curves of VGG16 and VGG19, we can see that as the complexity of the dataset increases, the plateau gradually disappears.
\end{itemize}

Since previous works~\cite{wang2018visualizing,DBLP:conf/eccv/ZeilerF14} have shown that shallow layers extract general features while deep layers learn more specifically. Hence, feature representations of the same category are more similar in deep layers than in shallow layers, which can be seen in \cref{fig:t-SNE visualization}. Therefore, the samples in the same community (category) tend to connect with each other, i.e., the modularity increases as the layer goes deeper. In this sense, the growth of modularity quantitatively reflects the process of class separation in feature representations inside the DNNs. Besides, the tendency of modularity is consistent with the observation of \cite{alain2016understanding}, which utilizes linear classifier probes to measure how suitable the feature representations at every layer are for classification. Compared to \cite{alain2016understanding} that requires training a linear classifier for each layer, our modularity provides a more convenient and effective tool to understand the dynamics of DNNs.

\begin{figure}[t]
\centering
 \subfigure[In adversarial scenarios.]
 {
  \begin{minipage}[c]{0.5\textwidth}
   \centering
   \includegraphics[width=0.99\textwidth]{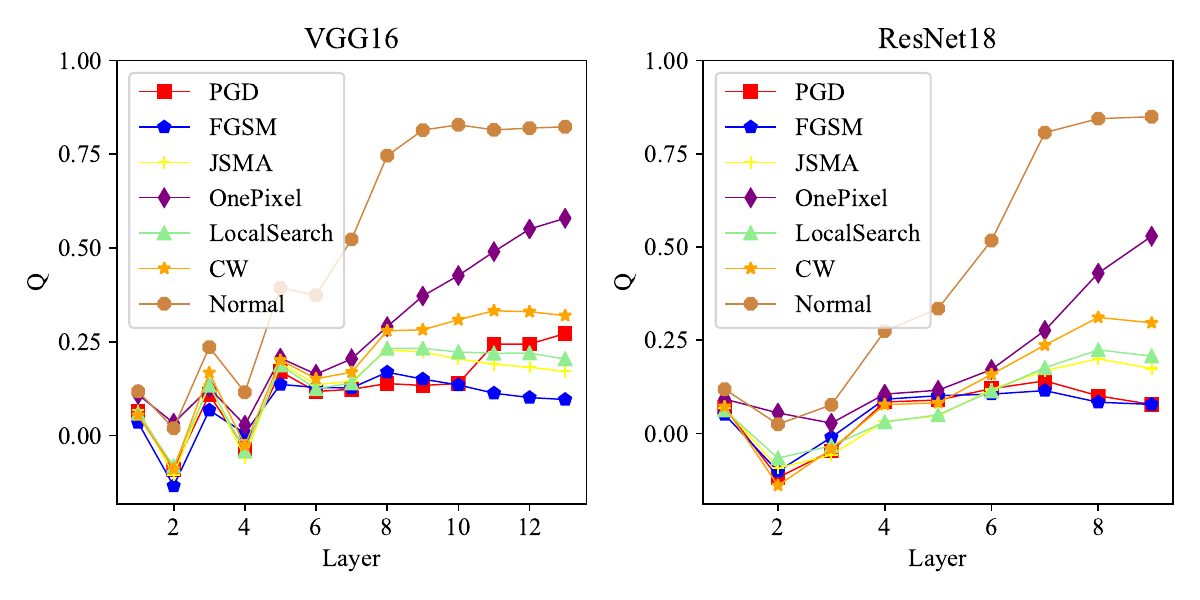}
  \end{minipage}%
  }%
 \subfigure[During training. Li denotes $i$-th layer.]
  {
  \begin{minipage}[c]{0.5\textwidth}
   \centering
   \includegraphics[width=0.99\textwidth]{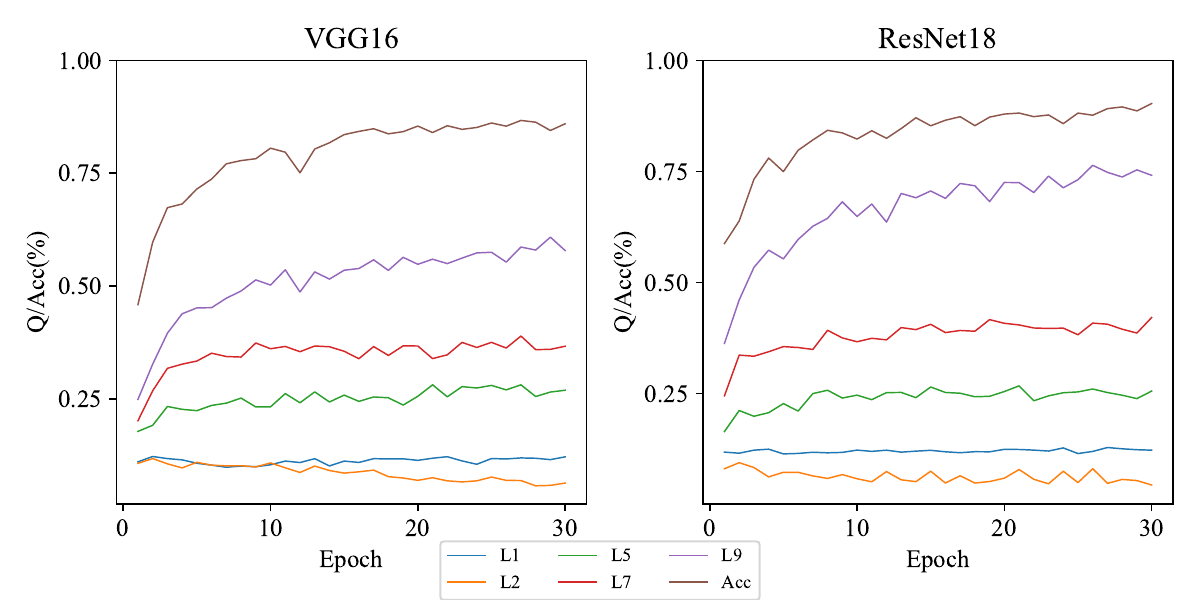}
  \end{minipage}
  }
\caption{Modularity curves in different scenarios.}
\label{fig:attack and trainng process}
\end{figure}

According to the second and third findings, we can draw a conclusion that the degradation and plateau arise when model complexity is great relative to the dataset. With the relative complexity of the model getting greater, a little bit of performance improvement costs nearly doubling the number of layers, e.g., ResNet18 and ResNet34 on CIFAR-10. Many successive layers are essentially performing the same operation over many times, refining feature representations just a little more each time instead of making a more fundamental change of feature representations. Hence, these features representations exhibit a similar degree of class separation, which explains the existence of the plateau and degradation in the modularity curves. To further explore the relationship between the degradation as well as plateau and model relative complexity, we use pre-trained ResNets in torchvision\footnote[1]{https://pytorch.org/vision/stable/index.html} to conduct experiments on ImageNet. From \cref{fig:ImageNet_modularity.pdf} we find that the tendency of modularity curves is almost consistent with the observation in \cref{modularity:VGGs and ResNets}(b). The main difference lies in the modularity curves of ResNet101 and ResNet152 trained on ImageNet50 exhibit the degradation and plateau in the middle layer, while those trained on ImageNet do not, which further confirms the conclusion we make. Additional experiment on exploring the evolution of communities is presented in Appendix A.1.

\begin{figure}[t]
\centering
 \subfigure[The influence of $N$ on the modularity.]
 {
  \begin{minipage}[c]{0.5\textwidth}
   \centering
   \includegraphics[width=0.99\textwidth]{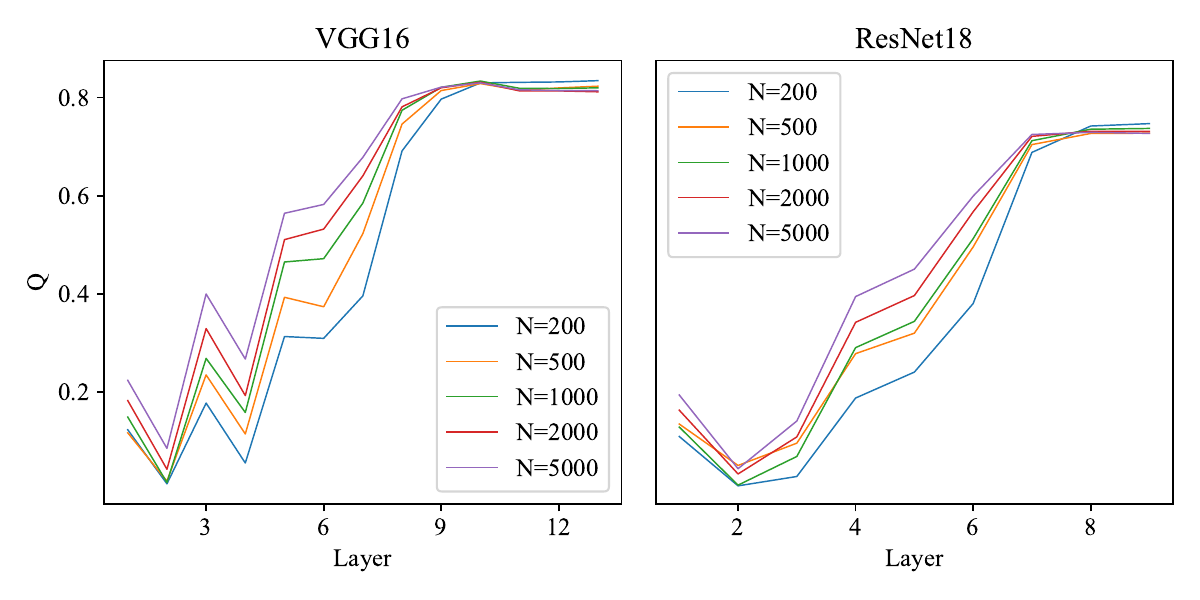}
  \end{minipage}%
  }%
 \subfigure[The influence of $k$ on the modularity.]
  {
  \begin{minipage}[c]{0.5\textwidth}
   \centering
   \includegraphics[width=0.99\textwidth]{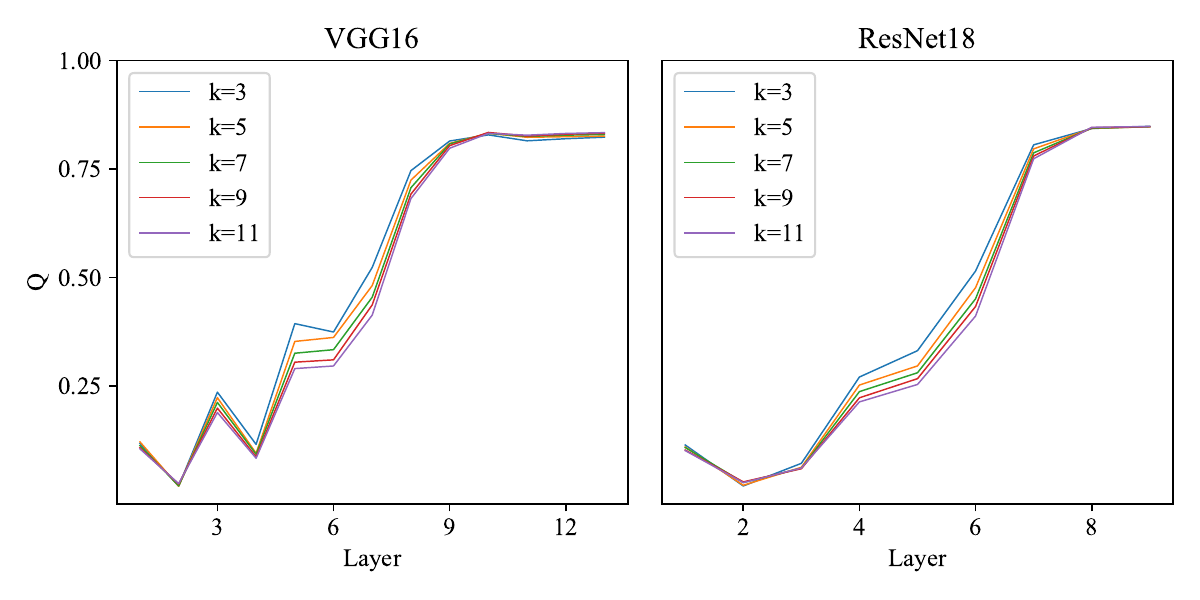}
  \end{minipage}
  }
\caption{Hyperparameter sensitivity analysis.}
\label{fig:k and n}
\end{figure}

\subsection{Modularity Curves of Adversarial Samples}
In \cref{Understanding the Evolution of Communities}, we discuss the modularity curves of normal samples. Here, we would like to explore the modularity curves of adversarial samples. Specifically, we first utilize FGSM \cite{DBLP:journals/corr/GoodfellowSS14}, PGD \cite{DBLP:conf/iclr/MadryMSTV18}, JSMA \cite{papernot2016limitations}, CW \cite{carlini2017towards}, OnePixel \cite{su2019one} and Local Search \cite{narodytska2016simple} to attack the pre-trained VGG16 and ResNet18 on CIFAR-10 for generating adversarial samples. To make it easier for others to reproduce our results, we utilize default parameter settings in AdverTorch \cite{ding2019advertorch} to obtain untargeted adversarial samples. The attack success rate is 100\% for each attack mode. Then we randomly choose 500 adversarial samples in each attack mode, set $k = 3$ to construct dynamic graphs and plot modularity curves. As shown in \cref{fig:attack and trainng process}(a), we can find that the modularity curve of adversarial samples can reach a smaller peak than normal samples, which can be interpreted as adversarial attacks blur the distinctions among various categories. 

\subsection{Modularity during Training Time} 
During the training process, the model is considered to learn valid feature representations. In order to explore the dynamics of DNNs in this process, we conduct experiments on ResNet18 and VGG16. Specifically, in the first 30 training epochs, we save the model file and test accuracy of every epoch. Then we construct a dynamic graph for each model and calculate the modularity of each snapshot in this dynamic graph. \cref{fig:attack and trainng process}(b) shows the modularity and accuracy curves on ResNet18 and VGG16, from which we find that the values of modularity in shallow layers nearly keep constant or small fluctuations. We believe that is because feature representations in shallow layers are general~\cite{wang2018visualizing,DBLP:conf/eccv/ZeilerF14}, samples in the same category do not cluster together. Hence, despite learning effective feature representations, the value of modularity does not increase significantly. Compared to shallow layers, feature representations in deep layers are more specific. With the improvement of model performance, the valid feature representations gradually learned by the model make intra-community connections tighter, which explains the modularity curves of deep layers are almost proportional to the accuracy curve. These phenomena may give some information about how the training evolves inside the DNNs and guide the intuition of researchers. Besides, we provide the experiment on randomly initialized models in Appendix A.2.

\subsection{Ablation Study} To provide further insight into the modularity, we conduct ablation studies to evaluate the hyperparameter sensitivity of the modularity. The batch size $N$ and the number of edges $k$ that each node connects with are two hyperparameters. 

\textbf{Whether batch size $N$ has a non-negligible impact on the modularity curve?} we set $N=200, 500, 1000, 2000, 5000$, $k = 3$ to repeat experiments on CIFAR-10 with pre-trained VGG16 and ResNet18. In \cref{fig:k and n}(a), we show that smaller $N$ has relatively lower modularity in early layers but $N$ has less influence on modularity in final layers. Generally speaking, different values of $N$ have the same tendency on modularity curves.

\textbf{Whether $k$ has a non-negligible impact on the modularity curve?} we set $N=500$, $k = 3, 5, 7, 9, 11$ to repeat the above experiments. \cref{fig:k and n}(b) shows that the different selections of $k$ have a negligible impact on modularity because the modularity curves almost overlap together. We observe the similar tendency even when $k$ is large (see Appendix A.3 for details)).

Hence, we can conclude that the modularity is reliable for hyperparameters.

\begin{figure}[t]
\centering
 \subfigure[Ours]
 {
  \begin{minipage}[c]{0.5\textwidth}
   \centering
   \includegraphics[width=0.9\textwidth]{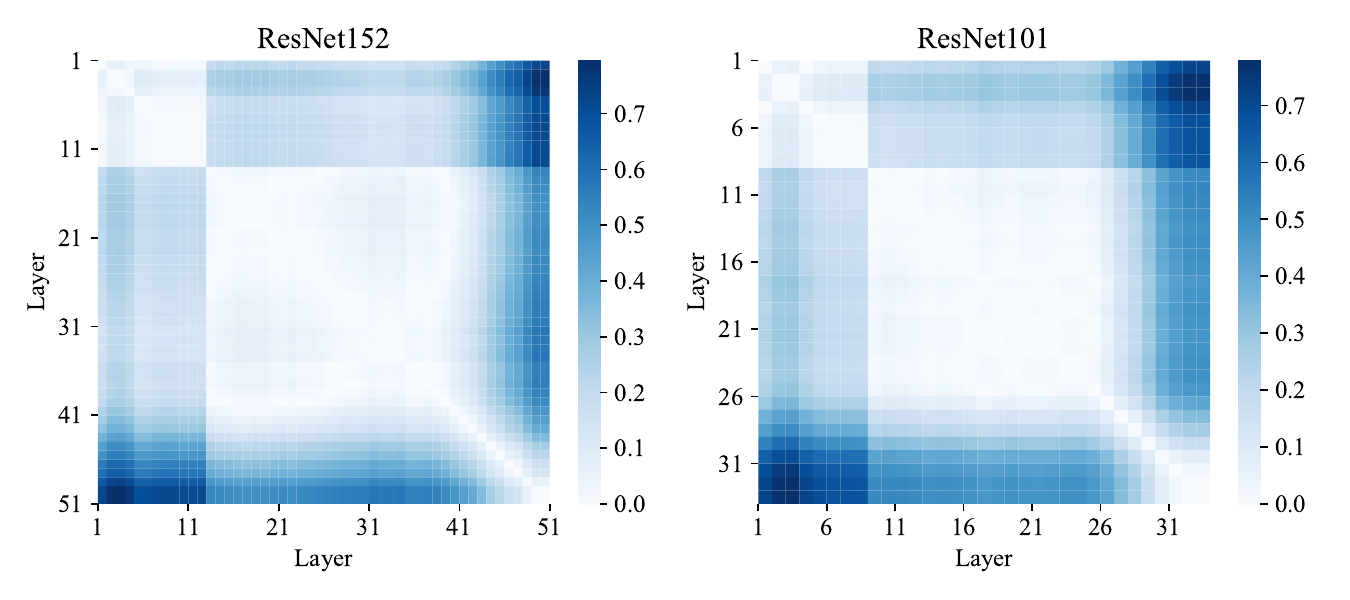}
  \end{minipage}%
  }%
 \subfigure[Linear CKA]
  {
  \begin{minipage}[c]{0.5\textwidth}
   \centering
   \includegraphics[width=0.9\textwidth]{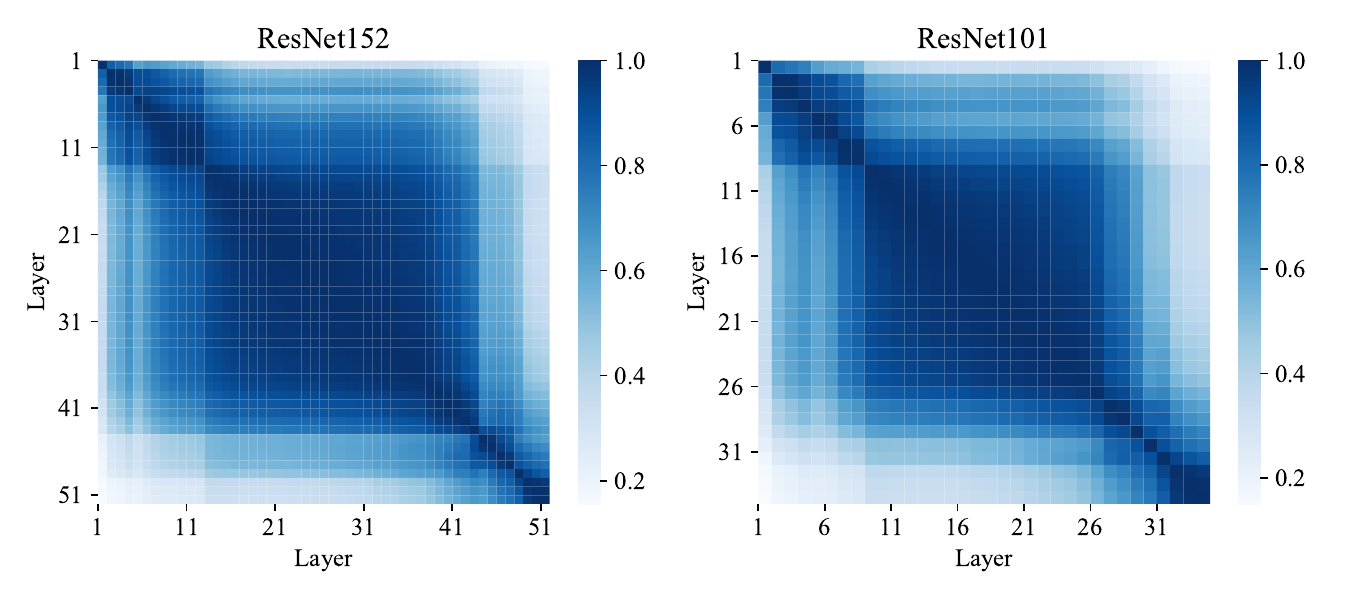}
  \end{minipage}
  }
\caption{Emergence of the block structure with different methods.}
\label{fig:modularity and CKA map}
\end{figure}

\section{Application Scenarios of Modularity}
\label{sec:Application Scenarios of Modularity}
In \cref{experiments}, we investigate the dynamics of DNNs, from which we gain some insights. On this basis, we provide two application scenarios for the modularity.

\subsection{Representing the Difference of Various Layers} 
\label{sec:Representing the Difference of Various Layers}
In addition to quantifying the degree of class separation of intermediate representations, modularity can also be used to represent the difference of various layers like previous works \cite{kornblith2019similarity,DBLP:conf/iclr/NguyenRK21,DBLP:conf/nips/RaghuGYS17,DBLP:conf/nips/MorcosRB18,DBLP:conf/nips/WangHGHW0H18,tang2020similarity,feng2020transferred}. Instead of directly calculating the similarity between two feature representations, we compute the difference of modularity of two layers because our modularity reflects the global attribute of the corresponding layer, i.e., the degree of class separation. Specifically, we calculate the difference matrix $D$, with its element defined as $\mathit{D}_{ij} = | \mathit{Q}_i - \mathit{Q}_j |$. $\mathit{Q}_i$ and $\mathit{Q}_j$ denote the value of modularity in $i$-th and $j$-th layer, respectively. We visualize the result as a heatmap, with the $x$ and $y$ axes representing the layers of the model. As shown in \cref{fig:modularity and CKA map}(a), the heatmap shows a block structure in representational difference, which arises because representations after residual connections are more different from representations inside ResNet blocks than other post-residual representations. Moreover, we also reproduce the result of linear CKA \cite{kornblith2019similarity} in \cref{fig:modularity and CKA map}(b), from which we can see the same block structure. Note that we measure the difference of various layers while \cite{kornblith2019similarity} focuses on the similarity, so the darker the color in \cref{fig:modularity and CKA map}(b), the less similar.

\subsection{Guiding Layer Pruning with Modularity}
\label{sec:Guiding Layer Pruning with Modularity}
In \cref{Understanding the Evolution of Communities}, we find that the degradation and plateau arise when model complexity is great relative to the dataset. Previous works have demonstrated that DNNs are redundant in depth \cite{DBLP:conf/bmvc/ZagoruykoK16} and overparameterized models exist many consecutive hidden layers that have highly similar feature representations \cite{DBLP:conf/iclr/NguyenRK21}. Informed by these conclusions, we wonder if the degradation and plateau offer an intuitive instruction in identifying the redundant layers. Hence, we assume that the plateau and degradation make no contribution or negative contribution to the model. In other words, these layers we consider are redundant and can be pruned with acceptable loss. To verify our assumption, we conduct systematic experiments on CIFAR-10, CIFAR-100 and ImageNet50.

\begin{figure*}[!t]
  \centering
   \includegraphics[width=0.9\linewidth]{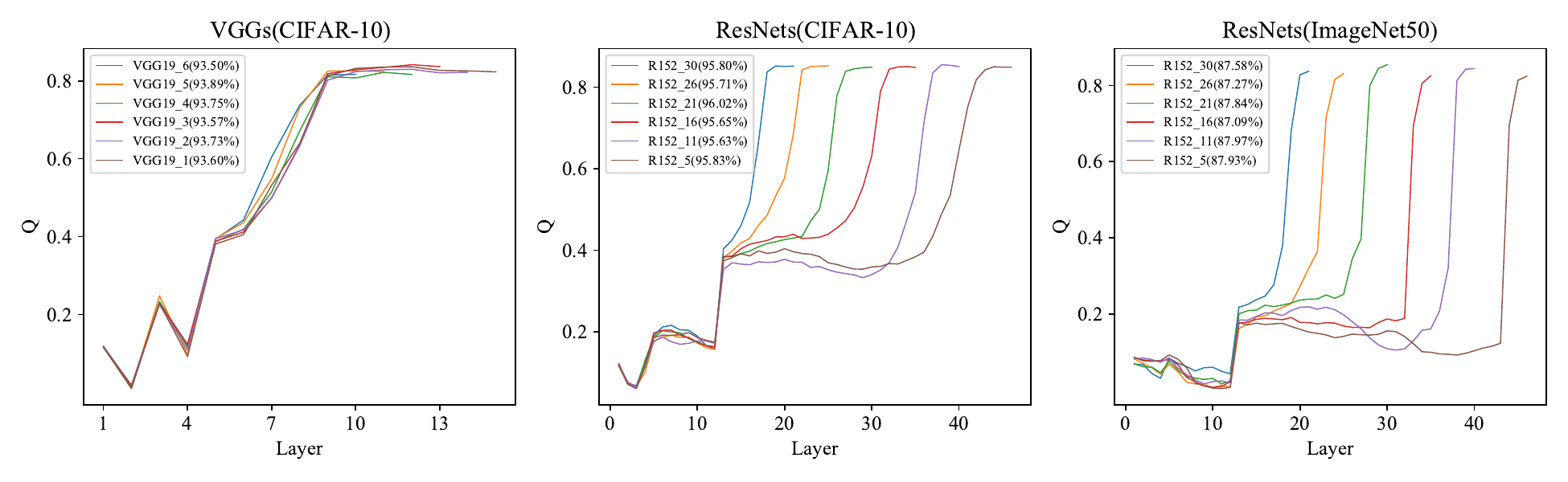}
   \caption{The modularity curves of VGGs and ResNets of different depths.}
   \label{fig:redundancy}
\end{figure*}

\begin{figure}[t]
\centering
 \subfigure[The modularity curves of VGGs.]
 {
  \begin{minipage}[c]{0.5\textwidth}
   \centering
   \includegraphics[width=0.9\textwidth]{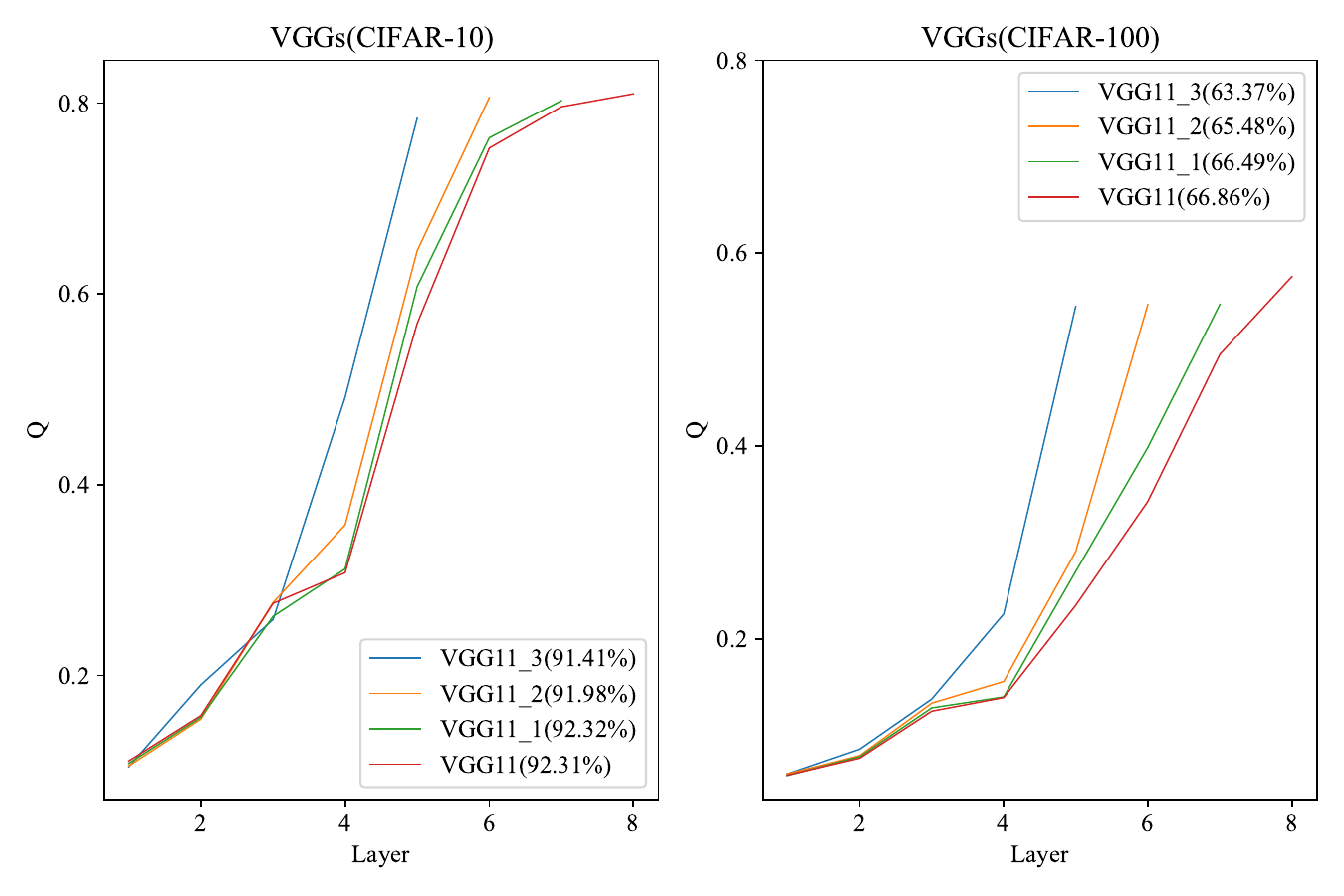}
  \end{minipage}%
  }%
 \subfigure[The correlation of $\Delta Q$ and $\Delta Acc$.]
  {
  \begin{minipage}[c]{0.5\textwidth}
   \centering
   \includegraphics[width=0.9\textwidth]{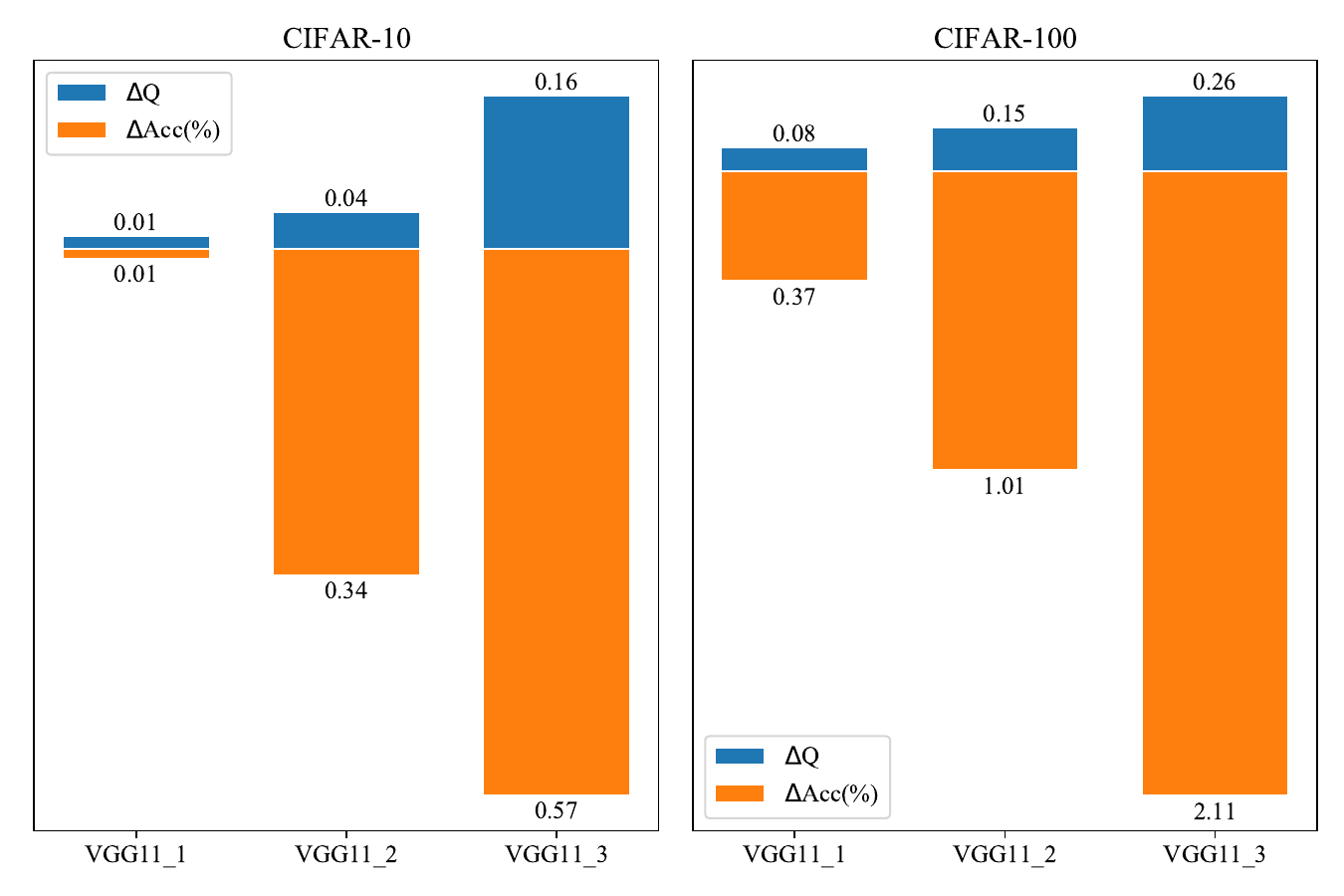}
  \end{minipage}
  }
\caption{Experiments on pruning the irredundant layers.}
\label{fig:line and dot}
\end{figure}

\textbf{Pruning the redundant layers.} Since the plateau mostly appears in the last few layers of VGG19, we remove the Conv-BN-ReLu in VGG19 one by one from back to front to obtain a series of variants. As for ResNet152, we remove the bottleneck building blocks from back to front in stage 3 (the degradation and plateau mostly emerge in stage 3) to get variant models. Detailed structures are shown in Appendix B. For example, VGG19\_1 denotes VGG19 prunes 1 layer. Then we finetune these variant models with the same hyperparameter setting as section 4.1. The left part of \cref{fig:redundancy} shows the results of VGGs on CIFAR-10, from which we find that the modularity curves of different VGG almost overlap together. The only difference between these modularity curves is whether the plateau emerges in the last few layers. Specifically, the modularity curve of VGG19\_6 does not have the plateau, while VGG19\_1 has the obvious plateau. Moreover, the plateau gradually disappears as the layer is removed one by one from back to front. Note that these variant models have similar performance, which proves that the plateau is indeed redundant. The middle part and right part of \cref{fig:redundancy} show the modularity curves of different ResNet on CIFAR-10 and ImageNet50. These modularity curves almost coincide in the shallow layers, while the plateau gradually narrows with the continuous removal of the middle layers.
Consequently, this strongly proves that the plateau can be pruned with minimal impact on performance.

\textbf{Pruning the irredundant layers.} In the previous paragraph, we verify that removing redundant layers will not affect the accuracy. Here, we wonder whether removing irredundant layers will result in a significant performance drop. Hence, we conduct further experiments on CIFAR-10 and CIFAR-100 with VGG11 (According to the modularity curves in \cref{modularity:VGGs and ResNets}, we think VGG11 is relatively irredundant on CIFAR-10 and CIFAR-100). We remove the Conv-BN-ReLu in VGG11 one by one from back to front to obtain a series of variants and finetune them. \cref{fig:line and dot}(a) exhibits the modularity curves of those variants, from which we can see that they can finally reach almost the same peak. Next we calculate the corresponding variation of accuracy brought about by pruning the layer. Specifically, we calculate the variation of modularity $\Delta Q$ of the final layer with $\Delta Q = \mathit{Q}_{i} - \mathit{Q}_{i-1}$, where $\mathit{Q}_{i}$ denotes the modularity of $i$-th layer that we want to prune. Finally, we calculate the variation of accuracy $\Delta Acc$ using $\Delta Acc = \mathit{Acc}_{i} - \mathit{Acc}_{j}$, where $\mathit{Acc}_{i}$ and $\mathit{Acc}_{j}$ represent the accuracy of the original model and pruned model, respectively. \cref{fig:line and dot}(b) shows the results, from which we can find that on CIFAR-10, removing the layer that modularity increases 0.01 results in nearly no influence (0.01\%) on model performance, while pruning the layer that has a 0.16 increment on modularity results in 0.57\% degradation. With the complexity of the dataset increasing, this gap becomes more obvious. On CIFAR-100, pruning the layer that modularity goes up 0.08 leads to a 0.37\% drop in accuracy, while a variation of 0.26 in modularity causes a variation of 2.11\% in accuracy. Hence, we draw a conclusion that the variation of accuracy is proportional to the variation of modularity, which means pruning irredundant layers will result in a more significant drop in performance than removing redundant layers. Besides, This phenomenon becomes more obvious as the complexity of the dataset increases.

\textbf{Practicality of layer pruning by modularity.} According to the above experimental results, we are able to conclude that modularity can be used to provide effective theoretical guidance for layer pruning. Here, we would like to evaluate its practicality. Specifically, we first plot the modularity curve of the original model, then we prune the layer where the curve drops, reaches a plateau or grows slowly, finally we finetune the new model. We adapt number of parameters and required Float Points Operations (denoted as FLOPs), to evaluate model size and computational requirement. We leverage a package in pytorch\cite{paszke2017automatic}, which terms thop\footnote[2]{https://github.com/Lyken17/pytorch-OpCounter} to calculate FLOPs and parameters. \cref{table:pruning resnet56} shows the performance of different layer pruning methods~\cite{chen2018shallowing,wang2019dbp} on ResNet56 for CIFAR. Compared with Chen et al, our method provides considerable better parameters and FLOPs reductions (43.00\% vs. 42.30\%, 60.30\% vs. 34.80\%), while yielding a higher accuracy (93.38\% vs. 93.29\%). Compared to DBP-0.5, our method shows more advantages in FLOPs reduction (60.30\% vs. 53.41\%), while maintaining a competitive accuracy (93.38\% vs. 93.39\%). 

According to the above experiments, we demonstrate the effectiveness and efficiency of layer pruning guided by modularity. 

\begin{table*}[t]
\centering
\caption{Pruning results of ResNet56 on CIFAR-10. PR is the pruning rate.}
\label{table:pruning resnet56}
\begin{tabular}{ccccc}
\hline
Method     & Top-1\% & Params(PR) & FLOPs(PR) \\ \hline
ResNet56   & 93.27   & 0\%        & 0\%       \\ 
Chen et al~\cite{chen2018shallowing} & 93.29   & 42.30\%    & 34.80\%   \\ 
DBP-0.5~\cite{wang2019dbp}    & 93.39   & /          & 53.41\%   \\ 
Ours       & 93.38   & 43.00\%    & 60.30\%   \\ \hline
\end{tabular}
\end{table*}

\section{Conclusion and Future Work}
In this study, through modeling the process of class separation from layer to layer as the evolution of communities in dynamic graphs, we provide a graph perspective for researchers to better understand the dynamics of DNNs. Then we develop modularity as a conceptual tool and apply it to various scenarios, e.g., the training process, standard and adversarial scenarios, to gain insights into the dynamics of DNNs. Extensive experiments show that modularity tends to rise as the layer goes deeper, which quantitatively reveals the process of class separation in intermediate layers. Moreover, the degradation and plateau arise when model complexity is great relative to the dataset. Through further analysis on the degradation and plateau at particular layers, we demonstrate that modularity can provide theoretical guidance for layer pruning. In addition to guiding layer pruning, modularity can also be used to represent the difference of various layers.

We hope the simplicity of our dynamic graph construction approach could facilitate more research ideas in interpreting DNNs from a graph perspective. Besides, we wish that the modularity presented in this paper can make a tiny step forward in the direction of neural network structure design, layer pruning and other potential applications. Recent work has shown that Vision Transformers can achieve superior performance on image classification tasks~\cite{liu2021swin,DBLP:conf/iclr/DosovitskiyB0WZ21}. In the future, we will further explore the dynamics of Visual Transformers.

~\\
\noindent \textbf{Acknowledgements.} This work was supported in part by the Key R\&D Program of Zhejiang under Grant 2022C01018, by the National Natural Science Foundation of China under Grants U21B2001, 61973273, 62072406, 11931015, U1803263, by the Zhejiang Provincial Natural Science Foundation of China under Grant LR19F030001, by the National Science Fund for Distinguished Young Scholars under Grant 62025602, by the Fok Ying-Tong Education Foundation, China under Grant 171105, and by the Tencent Foundation and XPLORER PRIZE. We also sincerely thank Jinhuan Wang, Zhuangzhi Chen and Shengbo Gong for their excellent suggestions.

\clearpage

\appendix
\renewcommand\thefigure{\Alph{section}\arabic{figure}} 
\renewcommand\thetable{\Alph{section}\arabic{table}}   

\title{Supplementary Material\\ Understanding the Dynamics of DNNs Using Graph Modularity} 

\titlerunning{Graph Modularity}

\author{Yao Lu\inst{1}\orcidlink{0000-0003-0655-7814}
\and
Wen Yang\inst{1}\orcidlink{0000-0002-8525-5672} \and
Yunzhe Zhang\inst{1}\orcidlink{0000-0003-2662-608X} \and Zuohui Chen\inst{1}\orcidlink{0000-0003-1806-6676} \and Jinyin Chen\inst{1}\orcidlink{0000-0002-7153-2755} \and \\ Qi Xuan\inst{1} \textsuperscript{\Letter}\orcidlink{0000-0002-6320-7012} \and Zhen Wang\inst{2} \textsuperscript{\Letter}\orcidlink{0000-0002-8182-2852} \and Xiaoniu Yang\inst{1,3}\orcidlink{0000-0003-3117-2211}}
\authorrunning{Y. Lu et al.}

\institute{Institute of Cyberspace Security, Zhejiang University of Technology, Hangzhou, 310023. China\\ \email{\{yaolu.zjut,czuohui\}@gmail.com}, \email{\{chenjinyin,xuanqi\}@zjut.edu.cn}, \email{wenyang.zjut@outlook.com}, \email{xsgxlz@live.cn} \and School of Artificial Intelligence, Optics and Electronics (iOPEN), Northwestern Polytechnical University, Xi’an 710072. China\\ \email{zhenwang0@gmail.com} \and Science and Technology on Communication Information Security Control Laboratory, Jiaxing 314033, China\\ \email{yxn2117@126.com}} 
\maketitle

\noindent This document supplements our paper by providing additional experiments and experimental details.

\section{Supplementary Experiments}
\subsection{Different Similarity indexes to Calculate the Similarity Matrix}
\begin{figure}[hbpt]
  \centering
  \includegraphics[width=0.99\linewidth]{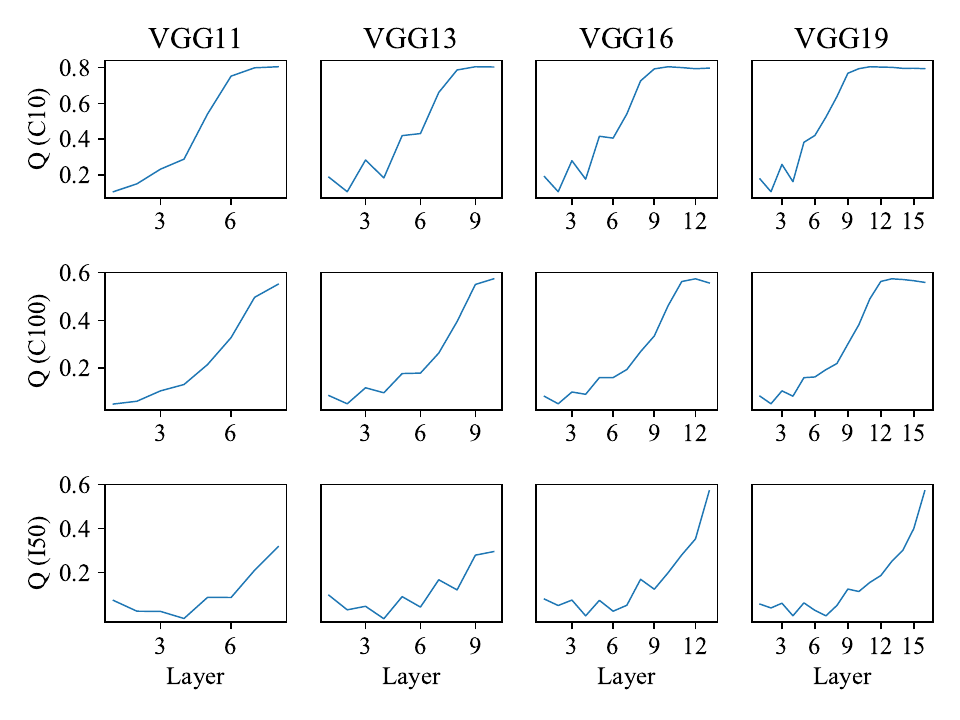}
  \caption{We utilize pearson correlation coefficient instead of cosine similarity to calculate the similarity matrix. We can find the modularity curves are almost same as those in the main text, which demonstrates that our modularity can be obtained by various indexes.}
  \label{fig:pearson_vgg}
\end{figure}

In this subsection, we replace cosine similarity with pearson correlation coefficient to calculate the similarity matrix. We can observe from \cref{fig:pearson_vgg} that the modularity curves are almost the same as those in the manuscript, demonstrating that various similarity indexes can be utilized to obtain the modularity.

\subsection{Modularity curves of Randomly Initialized Models}
\label{sec:Modularity curves of Randomly Initialized Models}
\begin{figure}[hbpt]
  \centering
  \includegraphics[width=0.99\linewidth]{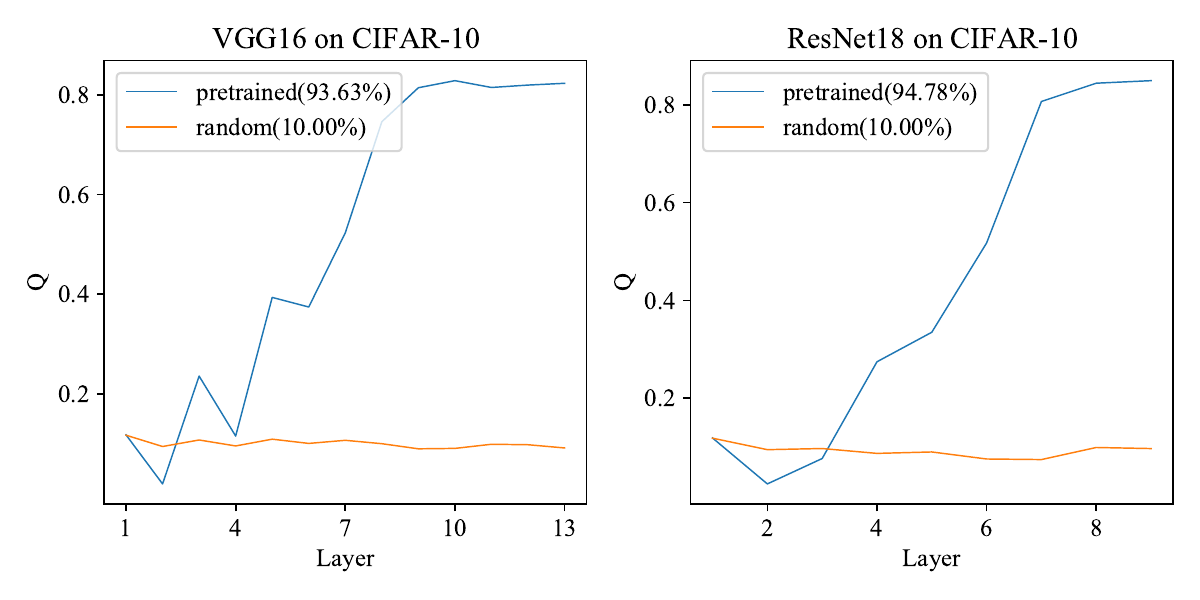}
  \caption{Modularity curves of well-optimized and randomly initialized models.}
  \label{fig:random}
\end{figure}

Having observed the upward trend of modularity emerging in pre-trained models, we next study whether this phenomenon exists in randomly initialized models. Hence, we conduct experiments with VGG16 and ResNet18 on CIFAR-10. In \cref{fig:random}, we show the modularity curves of randomly initialized models. Since the randomly initialized models can not learn useful feature representations, the connections among samples at each layer are completely random. Thus, the modularity nearly maintains constant in the randomly initialized models compared to pre-trained models.

\subsection{Ablation of $k$}
\label{sec:Ablation of k}
\begin{figure}[htbp]
  \centering
  \includegraphics[width=0.7\linewidth]{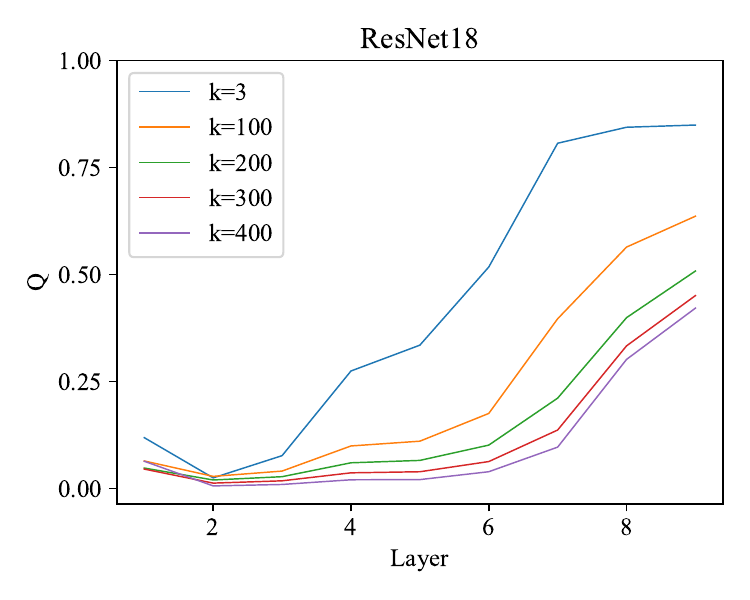}
  \caption{The influence of k on the modularity when k is large.}
  \label{fig:rebuttal}
\end{figure}

In Fig.7(b), the variability of $k$ is small, it is reasonable to question whether $k$ has a significant impact when $k$ is large ($k >> 11$). To answer this doubt, we set $N=500$, $k=100, 200, 300, 400$ to repeat the experiment. \cref{fig:rebuttal} shows the result, from which we find that different values of $k$ have a certain impact on modularity values but do not hurt the tendency of modularity curves. When $k$ is large, each node connects many unrelated nodes (i.e., nodes belonging to different communities), which weakens the intra-community connections but strengthens the inter-community connections, resulting in a decrease in the modularity value.

\section{Implementation Details}
\label{sec:Implementation Details}
\subsection{ImageNet50}
The 50 classes we randomly selected from ImageNet~\cite{DBLP:journals/ijcv/RussakovskyDSKS15} are shown as follows, you can also build the dataset yourself.
 
 n01443537 goldfish, Carassius auratus
 
 n01484850 great white shark, white shark, man-eater, man-eating shark, Carcharodon carcharias
 
 n01491361 tiger shark, Galeocerdo cuvieri
 
 n01494475 hammerhead, hammerhead shark
 
 n01496331 electric ray, crampfish, numbfish, torpedo
 
 n01498041 stingray
 
 n01514668 cock
 
 n01514859 hen
 
 n01518878 ostrich, Struthio camelus
 
 n01530575 brambling, Fringilla montifringilla
 
 n01531178 goldfinch, Carduelis carduelis
 
 n01532829 house finch, linnet, Carpodacus mexicanus
 
 n01534433 junco, snowbird
 
 n01537544 indigo bunting, indigo finch, indigo bird, Passerina cyanea
 
 n01558993 robin, American robin, Turdus migratorius
 
 n01560419 bulbul
 
 n01580077 jay
 
 n01582220 magpie
 
 n01592084 chickadee
 
 n01601694 water ouzel, dipper
 
 n01608432 kite
 
 n01614925 bald eagle, American eagle, Haliaeetus leucocephalus
 
 n01616318 vulture
 
 n01632777 axolotl, mud puppy, Ambystoma mexicanum
 
 n01667778 terrapin
 
 n01688243 frilled lizard, Chlamydosaurus kingi
 
 n01728920 ringneck snake, ring-necked snake, ring snake
 
 n01773157 black and gold garden spider, Argiope aurantia
 
 n01795545 black grouse
 
 n01847000 drake
 
 n07768694 pomegranate
 
 n07802026 hay
 
 n07831146 carbonara
 
 n07836838 chocolate sauce, chocolate syrup
 
 n07860988 dough
 
 n07871810 meat loaf, meatloaf
 
 n07873807 pizza, pizza pie
 
 n09193705 alp
 
 n09229709 bubble
 
 n09246464 cliff, drop, drop-off
 
 n09256479 coral reef
 
 n09288635 geyser
 
 n09332890 lakeside, lakeshore
 
 n09399592 promontory, headland, head, foreland
 
 n09421951 sandbar, sand bar
 
 n09428293 seashore, coast, seacoast, sea-coast
 
 n09468604 valley, vale
 
 n09472597 volcano
 
 n09835506 ballplayer, baseball player
 
 n10148035 groom, bridegroom
 
\subsection{Statistics of Models}
The statistics of pre-trained models used in this paper are shown in \cref{table:acc}. Moreover, in order to better investigate the redundant layers in models. We design a series of variants. The model structures of these variants are shown in Table B2-B5.

\begin{table}[]
 \centering
  \caption{The statistics of pre-trained models.}
\label{table:acc}
\begin{tabular}{c|cccc|ccccc}
\hline
\multirow{2}{*}{Dataset} & \multicolumn{4}{c|}{VGGs}             & \multicolumn{5}{c}{ResNets}                    \\ \cline{2-10} 
                         & V11     & V13     & V16     & V19     & R18     & R34     & R50     & R101    & R152    \\ \hline
CIFAR-10                 & 92.11\% & 93.68\% & 93.63\% & 93.36\% & 94.78\% & 95.14\% & 95.38\% & 95.18\% & 95.32\% \\
CIFAR-100                & 66.87\% & 70.19\% & 72.34\% & 72.63\% & 76.87\% & 76.69\% & 78.03\% & 78.21\% & 78.64\% \\
ImageNet50               & 83.97\% & 85.87\% & 85.04\% & 87.32\% & 82.55\% & 82.99\% & 85.04\% & 87.32\% & 87.89\% \\ \hline
\end{tabular}
\end{table}

\begin{table*}
\label{tab:table1}
\centering
\caption{The model structures of variants of VGG19 on CIFAR-10.}
\begin{tabular}{|c|c|c|c|c|c|c|}
\hline
VGG19\_6 &
  VGG19\_5 &
 VGG19\_4 &
  VGG19\_3 &
  VGG19\_2 &
  VGG19\_1 &
  VGG19 \\ \hline
\multicolumn{7}{|c|}{input (32 × 32 RGB images)} \\ \hline
\begin{tabular}[c]{@{}c@{}}conv3-64\\ conv3-64\end{tabular} &
  \begin{tabular}[c]{@{}c@{}}conv3-64\\ conv3-64\end{tabular} &
  \begin{tabular}[c]{@{}c@{}}conv3-64\\ conv3-64\end{tabular} &
  \begin{tabular}[c]{@{}c@{}}conv3-64\\ conv3-64\end{tabular} &
  \begin{tabular}[c]{@{}c@{}}conv3-64\\ conv3-64\end{tabular} &
  \begin{tabular}[c]{@{}c@{}}conv3-64\\ conv3-64\end{tabular} &
  \begin{tabular}[c]{@{}c@{}}conv3-64\\ conv3-64\end{tabular} \\ \hline
\multicolumn{7}{|c|}{maxpool} \\ \hline
\begin{tabular}[c]{@{}c@{}}conv3-128\\ conv3-128\end{tabular} &
  \begin{tabular}[c]{@{}c@{}}conv3-128\\ conv3-128\end{tabular} &
  \begin{tabular}[c]{@{}c@{}}conv3-128\\ conv3-128\end{tabular} &
  \begin{tabular}[c]{@{}c@{}}conv3-128\\ conv3-128\end{tabular} &
  \begin{tabular}[c]{@{}c@{}}conv3-128\\ conv3-128\end{tabular} &
  \begin{tabular}[c]{@{}c@{}}conv3-128\\ conv3-128\end{tabular} &
  \begin{tabular}[c]{@{}c@{}}conv3-128\\ conv3-128\end{tabular} \\ \hline
\multicolumn{7}{|c|}{maxpool} \\ \hline
\begin{tabular}[c]{@{}c@{}}conv3-256\\ conv3-256\\ conv3-256\\ conv3-256\end{tabular} &
  \begin{tabular}[c]{@{}c@{}}conv3-256\\ conv3-256\\ conv3-256\\ conv3-256\end{tabular} &
  \begin{tabular}[c]{@{}c@{}}conv3-256\\ conv3-256\\ conv3-256\\ conv3-256\end{tabular} &
  \begin{tabular}[c]{@{}c@{}}conv3-256\\ conv3-256\\ conv3-256\\ conv3-256\end{tabular} &
  \begin{tabular}[c]{@{}c@{}}conv3-256\\ conv3-256\\ conv3-256\\ conv3-256\end{tabular} &
  \begin{tabular}[c]{@{}c@{}}conv3-256\\ conv3-256\\ conv3-256\\ conv3-256\end{tabular} &
  \begin{tabular}[c]{@{}c@{}}conv3-256\\ conv3-256\\ conv3-256\\ conv3-256\end{tabular} \\ \hline
\multicolumn{7}{|c|}{maxpool} \\ \hline
conv3-512 &
  \begin{tabular}[c]{@{}c@{}}conv3-512\\ conv3-512\end{tabular} &
  \begin{tabular}[c]{@{}c@{}}conv3-512\\ conv3-512\\ conv3-512\end{tabular} &
  \begin{tabular}[c]{@{}c@{}}conv3-512\\ conv3-512\\ conv3-512\\ conv3-512\end{tabular} &
  \begin{tabular}[c]{@{}c@{}}conv3-512\\ conv3-512\\ conv3-512\\ conv3-512\end{tabular} &
  \begin{tabular}[c]{@{}c@{}}conv3-512\\ conv3-512\\ conv3-512\\ conv3-512\end{tabular} &
  \begin{tabular}[c]{@{}c@{}}conv3-512\\ conv3-512\\ conv3-512\\ conv3-512\end{tabular} \\ \hline
\multicolumn{7}{|c|}{maxpool} \\ \hline
conv3-512 &
  conv3-512 &
  conv3-512 &
  conv3-512 &
  \begin{tabular}[c]{@{}c@{}}conv3-512\\ conv3-512\end{tabular} &
  \begin{tabular}[c]{@{}c@{}}conv3-512\\ conv3-512\\ conv3-512\end{tabular} &
  \begin{tabular}[c]{@{}c@{}}conv3-512\\ conv3-512\\ conv3-512\\ conv3-512\end{tabular} \\ \hline
\multicolumn{7}{|c|}{maxpool} \\ \hline
\multicolumn{7}{|c|}{FC-512} \\ \hline
\multicolumn{7}{|c|}{FC-512} \\ \hline
\multicolumn{7}{|c|}{FC-10} \\ \hline
\multicolumn{7}{|c|}{soft-max} \\ \hline
\end{tabular}
\end{table*}

\begin{table*}[]
\label{tab:table2}
\centering
\caption{The model structures of variants of ResNet152 on CIFAR-10.}
\resizebox{\textwidth}{33mm}{
    \begin{tabular}{|c|c|c|c|c|c|c|c|c|c|c|c|c|c|}
    \hline
    Layer &
      Size &
      \multicolumn{2}{c|}{Res152\_30} &
      \multicolumn{2}{c|}{Res152\_26} &
      \multicolumn{2}{c|}{Res152\_21} &
      \multicolumn{2}{c|}{Res152\_16} &
      \multicolumn{2}{c|}{Res152\_11} &
      \multicolumn{2}{c|}{Res152\_5} \\ \hline
    conv1 &
      32×32 &
      \multicolumn{12}{c|}{3×3, 64, stride 1} \\ \hline
    \multirow{2}{*}{conv2\_x} &
      \multirow{2}{*}{32×32} &
      \multicolumn{12}{c|}{                } \\ \cline{3-14} 
     &
      &
      \begin{tabular}[c]{@{}c@{}}1×1, 64\\ 3×3, 64\\ 1×1, 256\end{tabular} &
      3 &
      \begin{tabular}[c]{@{}c@{}}1×1, 64\\ 3×3, 64\\ 1×1, 256\end{tabular} &
      3 &
      \begin{tabular}[c]{@{}c@{}}1×1, 64\\ 3×3, 64\\ 1×1, 256\end{tabular} &
      3 &
      \begin{tabular}[c]{@{}c@{}}1×1, 64\\ 3×3, 64\\ 1×1, 256\end{tabular} &
      3 &
      \begin{tabular}[c]{@{}c@{}}1×1, 64\\ 3×3, 64\\ 1×1, 256\end{tabular} &
      3 &
      \begin{tabular}[c]{@{}c@{}}1×1, 64\\ 3×3, 64\\ 1×1, 256\end{tabular} &
      3 \\ \hline
    conv3\_x &
      16×16 &
      \begin{tabular}[c]{@{}c@{}}1×1, 128\\ 3×3, 128\\ 1×1, 512\end{tabular} &
      8 &
      \begin{tabular}[c]{@{}c@{}}1×1, 128\\ 3×3, 128\\ 1×1, 512\end{tabular} &
      8 &
      \begin{tabular}[c]{@{}c@{}}1×1, 128\\ 3×3, 128\\ 1×1, 512\end{tabular} &
      8 &
      \begin{tabular}[c]{@{}c@{}}1×1, 128\\ 3×3, 128\\ 1×1, 512\end{tabular} &
      8 &
      \begin{tabular}[c]{@{}c@{}}1×1, 128\\ 3×3, 128\\ 1×1, 512\end{tabular} &
      8 &
      \begin{tabular}[c]{@{}c@{}}1×1, 128\\ 3×3, 128\\ 1×1, 512\end{tabular} &
      8 \\ \hline
    conv4\_x &
      8×8 &
      \begin{tabular}[c]{@{}c@{}}1×1, 256\\ 3×3, 256\\ 1×1, 1024\end{tabular} &
      6 &
      \begin{tabular}[c]{@{}c@{}}1×1, 256\\ 3×3, 256\\ 1×1, 1024\end{tabular} &
      10 &
      \begin{tabular}[c]{@{}c@{}}1×1, 256\\ 3×3, 256\\ 1×1, 1024\end{tabular} &
      15 &
      \begin{tabular}[c]{@{}c@{}}1×1, 256\\ 3×3, 256\\ 1×1, 1024\end{tabular} &
      20 &
      \begin{tabular}[c]{@{}c@{}}1×1, 256\\ 3×3, 256\\ 1×1, 1024\end{tabular} &
      25 &
      \begin{tabular}[c]{@{}c@{}}1×1, 256\\ 3×3, 256\\ 1×1, 1024\end{tabular} &
      31 \\ \hline
    conv5\_x &
      4×4 &
      \begin{tabular}[c]{@{}c@{}}1×1, 512\\ 3×3, 512\\ 1×1, 2048\end{tabular} &
      3 &
      \begin{tabular}[c]{@{}c@{}}1×1, 512\\ 3×3, 512\\ 1×1, 2048\end{tabular} &
      3 &
      \begin{tabular}[c]{@{}c@{}}1×1, 512\\ 3×3, 512\\ 1×1, 2048\end{tabular} &
      3 &
      \begin{tabular}[c]{@{}c@{}}1×1, 512\\ 3×3, 512\\ 1×1, 2048\end{tabular} &
      3 &
      \begin{tabular}[c]{@{}c@{}}1×1, 512\\ 3×3, 512\\ 1×1, 2048\end{tabular} &
      3 &
      \begin{tabular}[c]{@{}c@{}}1×1, 512\\ 3×3, 512\\ 1×1, 2048\end{tabular} &
      3 \\ \hline
     &
      1×1 &
      \multicolumn{12}{c|}{adaptive average pool, 10-d fc, softmax} \\ \hline
    \end{tabular}}
\end{table*}

\begin{table*}[]
\label{tab:tabel3}
\centering
\caption{The model structures of variants of ResNet152 on ImageNet50.}
\resizebox{\textwidth}{33mm}{
    \begin{tabular}{|c|c|c|c|c|c|c|c|c|c|c|c|c|c|}
    \hline
    Layer &
      Size &
      \multicolumn{2}{c|}{Res152\_30} &
      \multicolumn{2}{c|}{Res152\_26} &
      \multicolumn{2}{c|}{Res152\_21} &
      \multicolumn{2}{c|}{Res152\_16} &
      \multicolumn{2}{c|}{Res152\_11} &
      \multicolumn{2}{c|}{Res152\_5} \\ \hline
    conv1 &
      112×112 &
      \multicolumn{12}{c|}{7×7, 64, stride 2} \\ \hline
    \multirow{2}{*}{conv2\_x} &
      \multirow{2}{*}{56×56} &
      \multicolumn{12}{c|}{3×3 max pool, stride 2} \\ \cline{3-14} 
     &
      &
      \begin{tabular}[c]{@{}c@{}}1×1, 64\\ 3×3, 64\\ 1×1, 256\end{tabular} &
      3 &
      \begin{tabular}[c]{@{}c@{}}1×1, 64\\ 3×3, 64\\ 1×1, 256\end{tabular} &
      3 &
      \begin{tabular}[c]{@{}c@{}}1×1, 64\\ 3×3, 64\\ 1×1, 256\end{tabular} &
      3 &
      \begin{tabular}[c]{@{}c@{}}1×1, 64\\ 3×3, 64\\ 1×1, 256\end{tabular} &
      3 &
      \begin{tabular}[c]{@{}c@{}}1×1, 64\\ 3×3, 64\\ 1×1, 256\end{tabular} &
      3 &
      \begin{tabular}[c]{@{}c@{}}1×1, 64\\ 3×3, 64\\ 1×1, 256\end{tabular} &
      3 \\ \hline
    conv3\_x &
      28×28 &
      \begin{tabular}[c]{@{}c@{}}1×1, 128\\ 3×3, 128\\ 1×1, 512\end{tabular} &
      8 &
      \begin{tabular}[c]{@{}c@{}}1×1, 128\\ 3×3, 128\\ 1×1, 512\end{tabular} &
      8 &
      \begin{tabular}[c]{@{}c@{}}1×1, 128\\ 3×3, 128\\ 1×1, 512\end{tabular} &
      8 &
      \begin{tabular}[c]{@{}c@{}}1×1, 128\\ 3×3, 128\\ 1×1, 512\end{tabular} &
      8 &
      \begin{tabular}[c]{@{}c@{}}1×1, 128\\ 3×3, 128\\ 1×1, 512\end{tabular} &
      8 &
      \begin{tabular}[c]{@{}c@{}}1×1, 128\\ 3×3, 128\\ 1×1, 512\end{tabular} &
      8 \\ \hline
    conv4\_x &
      14×14 &
      \begin{tabular}[c]{@{}c@{}}1×1, 256\\ 3×3, 256\\ 1×1, 1024\end{tabular} &
      6 &
      \begin{tabular}[c]{@{}c@{}}1×1, 256\\ 3×3, 256\\ 1×1, 1024\end{tabular} &
      10 &
      \begin{tabular}[c]{@{}c@{}}1×1, 256\\ 3×3, 256\\ 1×1, 1024\end{tabular} &
      15 &
      \begin{tabular}[c]{@{}c@{}}1×1, 256\\ 3×3, 256\\ 1×1, 1024\end{tabular} &
      20 &
      \begin{tabular}[c]{@{}c@{}}1×1, 256\\ 3×3, 256\\ 1×1, 1024\end{tabular} &
      25 &
      \begin{tabular}[c]{@{}c@{}}1×1, 256\\ 3×3, 256\\ 1×1, 1024\end{tabular} &
      31 \\ \hline
    conv5\_x &
      7×7 &
      \begin{tabular}[c]{@{}c@{}}1×1, 512\\ 3×3, 512\\ 1×1, 2048\end{tabular} &
      3 &
      \begin{tabular}[c]{@{}c@{}}1×1, 512\\ 3×3, 512\\ 1×1, 2048\end{tabular} &
      3 &
      \begin{tabular}[c]{@{}c@{}}1×1, 512\\ 3×3, 512\\ 1×1, 2048\end{tabular} &
      3 &
      \begin{tabular}[c]{@{}c@{}}1×1, 512\\ 3×3, 512\\ 1×1, 2048\end{tabular} &
      3 &
      \begin{tabular}[c]{@{}c@{}}1×1, 512\\ 3×3, 512\\ 1×1, 2048\end{tabular} &
      3 &
      \begin{tabular}[c]{@{}c@{}}1×1, 512\\ 3×3, 512\\ 1×1, 2048\end{tabular} &
      3 \\ \hline
     &
      1×1 &
      \multicolumn{12}{c|}{adaptive average pool, 50-d fc, softmax} \\ \hline
    \end{tabular}}
\end{table*}

\begin{table*}
\label{tab:table4}
\centering
\caption{The model structures of variants of VGG11 on CIFAR-10 and CIFAR-100.}
\begin{tabular}{|c|c|c|c|}
\hline
  VGG11\_3 &
  VGG11\_2 &
  VGG11\_1 &
  VGG11 \\ \hline
\multicolumn{4}{|c|}{input (32 × 32 RGB images)} \\ \hline
  \begin{tabular}[c]{@{}c@{}}conv3-64\end{tabular} &
  \begin{tabular}[c]{@{}c@{}}conv3-64\end{tabular} &
  \begin{tabular}[c]{@{}c@{}}conv3-64\end{tabular} &
  \begin{tabular}[c]{@{}c@{}}conv3-64\end{tabular} \\ \hline
\multicolumn{4}{|c|}{maxpool} \\ \hline
  \begin{tabular}[c]{@{}c@{}}conv3-128\end{tabular} &
  \begin{tabular}[c]{@{}c@{}}conv3-128\end{tabular} &
  \begin{tabular}[c]{@{}c@{}}conv3-128\end{tabular} &
  \begin{tabular}[c]{@{}c@{}}conv3-128\end{tabular} \\ \hline
\multicolumn{4}{|c|}{maxpool} \\ \hline
  \begin{tabular}[c]{@{}c@{}}conv3-256\end{tabular} &
  \begin{tabular}[c]{@{}c@{}}conv3-256\\ conv3-256\end{tabular} &
  \begin{tabular}[c]{@{}c@{}}conv3-256\\ conv3-256\end{tabular} &
  \begin{tabular}[c]{@{}c@{}}conv3-256\\ conv3-256\end{tabular} \\ \hline
\multicolumn{4}{|c|}{maxpool} \\ \hline
conv3-512 &
  \begin{tabular}[c]{@{}c@{}}conv3-512\end{tabular} &
  \begin{tabular}[c]{@{}c@{}}conv3-512\\ conv3-512\end{tabular} &
  \begin{tabular}[c]{@{}c@{}}conv3-512\\ conv3-512\end{tabular} \\ \hline
\multicolumn{4}{|c|}{maxpool} \\ \hline
  conv3-512 &
  \begin{tabular}[c]{@{}c@{}}conv3-512\end{tabular} &
  \begin{tabular}[c]{@{}c@{}}conv3-512\end{tabular} &
  \begin{tabular}[c]{@{}c@{}}conv3-512\\ conv3-512\end{tabular} \\ \hline
\multicolumn{4}{|c|}{maxpool} \\ \hline
\multicolumn{4}{|c|}{FC-512} \\ \hline
\multicolumn{4}{|c|}{FC-512} \\ \hline
\multicolumn{4}{|c|}{FC-10} \\ \hline
\multicolumn{4}{|c|}{soft-max} \\ \hline
\end{tabular}
\end{table*}

\end{document}